%% file: iclr2027_conference.tex
\documentclass{article} % For LaTeX2e
\usepackage{iclr2027_conference,times}

\input{math_commands.tex}

\usepackage{hyperref}
\usepackage{url}
\usepackage{algorithm}
\usepackage{algorithmic}
\usepackage{graphicx}
\usepackage{subcaption}
\usepackage{booktabs}

\title{Gated Target Propagation for Compositional Generalization in Continual Learning}

\newcommand{\shiftguillaume}[1]{%
  \makebox[0pt][l]{\hspace*{8mm}#1}\phantom{#1}%
}

\author{%
Abdel Mfougouon Njupoun\\
Université de Montréal\\
Mila Quebec AI Institute\\
\texttt{abdel.njupoun@mila.quebec}\\
\And
Colin Bredenberg\\
University of Oregon\\
Mila Quebec AI Institute\\
\texttt{colin.bredenberg@mila.quebec}\\
\AND
Blake Aaron Richards\\
McGill University\\
Mila Quebec AI Institute\\
\texttt{blake.richards@mila.quebec}\\
\And
\shiftguillaume{Guillaume Lajoie}\\
\shiftguillaume{Université de Montréal}\\
\shiftguillaume{Mila Quebec AI Institute}\\
\shiftguillaume{\texttt{guillaume.lajoie@mila.quebec}}%
}

\iclrfinalcopy % Uncomment for camera-ready version, but NOT for submission.
\begin{document}

\maketitle

\begin{abstract}
Continual learning is typically framed as acquiring new knowledge without catastrophically forgetting previous tasks. However, a flexible continual learner should also be able to reuse and recombine previously acquired knowledge to rapidly solve novel task compositions. We introduce Gated Target Propagation (GaTaP), a continual learning algorithm in which task-specific gating variables---learned through a closed-form inner loop update---selectively suppress or enhance network modules. Network parameters are learned in a slower timescale outer loop, using the same local difference target propagation error signal as is used for adapting gating variables. We provide tractable experiments on class-incremental learning scenarios for both multilayer perceptron and convolutional network architectures. We show strong performance retention on previously learned tasks, as well as compositional generalization to unseen tasks, achieved through few-shot gain adaptation at inference. We analyze learned gating patterns and find that related tasks exhibit similar gating patterns, suggesting that inferred gates capture meaningful, reusable task structure. Overall, GaTaP provides a powerful framework for jointly ameliorating catastrophic forgetting and enabling few-shot compositional generalization in neural network models.
\end{abstract}

\section{Introduction}
Continual learning aims to sequentially train models on new experiences while preserving knowledge acquired from previous ones. Neural networks, however, often suffer from \emph{catastrophic forgetting}: learning a new task can overwrite parameters that are important for earlier tasks, causing their performance to deteriorate~\citep{mccloskey1989catastrophic,french1999catastrophic,goodfellow2013empirical}. A large body of work addresses this problem through regularization~\citep{kirkpatrick2017overcoming,zenke2017continual}, replay~\citep{lopez2017gradient,rebuffi2017icarl,buzzega2020dark}, or architectural mechanisms that isolate task-specific parameters~\citep{rusu2016progressive,fernando2017pathnet,mallya2018packnet,serra2018overcoming, veniat2021efficientcontinuallearningmodular, rajasegaran2020adaptiverandompathselection}. While these approaches can improve retention, continual learning within these studies is typically evaluated only through the ability to remember previously learned tasks.

We argue that remembering alone is not sufficient for flexible continual learning. A capable learner should also be able to \emph{reuse and recombine} previously acquired knowledge to solve novel tasks. For example, after learning to discriminate between MNIST~\citep{lecun2010mnist} digits 3 and 4, and between 5 and 6, a model should ideally be able to adapt to unseen combinations such as $3$ vs.\ $6$ or $4$ vs.\ $5$ without relearning the underlying representations. This connects continual learning with \emph{compositional generalization}: the ability to solve novel combinations of previously encountered task components~\citep{lake2017generalization,elmozninoFormalTheoryRepresentational2025}. From this perspective, continual learning should not only preserve knowledge, but organize it into reusable computational components that can be rapidly recombined.

We introduce \textbf{GaTaP} (\textbf{Ga}ted \textbf{Ta}rget \textbf{P}ropagation), a learning framework designed to jointly address catastrophic forgetting and compositional generalization. GaTaP represents shared knowledge through a collection of subnetworks whose contributions are modulated by task-specific multiplicative gain variables, denoted by $\Gamma$. These gains act as soft gates that activate or inactivate subnetworks depending on the current task. The same shared weights can therefore support different tasks through different configurations of $\Gamma$, providing a mechanism for selective plasticity and knowledge reuse.

Training GaTaP networks is a form of supervised meta-learning. For a given task, an inner optimization loop selects a gain vector $\Gamma$  that gates subnetworks (fixed for inner loops), while an outer loop optimization updates subnetwork parameters for an ensemble of tasks, each with their own $\Gamma$. This separates \emph{fast task adaptation} from \emph{slow, shared-weight learning} through a  bilevel optimization scheme analogous to the inference and parameter-update steps of variational expectation maximization (VEM)~\citep{bernardo2003variational}. Crucially, optimization is carried out with difference target propagation~\citep{lee2015difference}. Rather than propagating an output gradient through the entire network, target propagation infers local target states for individual layers, providing a layer-wise credit-assignment mechanism that is local~\citep{bredenberg2023formalizing}, compatible with gated modular computation, and aligned with biologically plausible learning~\citep{bengio2014auto}. The use of this local learning rule has a distinctive implication for the inner loop: {\bf learning optimal gain parameters for a new task reduces to a series of least square problems which can be solved in closed form, enabling adaptation at inference without gradient descent}.

The compositional generalization capabilities of GaTaP thus manifest at \emph{inference time}, once the outer loop training is complete and all shared weights are frozen.   In this setting, adaptation to either previously learned or novel tasks is performed solely by inferring a new gain configuration $\Gamma$ from a small support set (a simple closed-form evaluation). Consequently, success on an unseen task directly tests whether the model has acquired reusable computational components: no modification of the shared representation is allowed during adaptation.

Our main contributions are:
\begin{itemize}
    \item We formulate continual learning as a joint problem of \emph{retention and composition}, requiring learned knowledge to remain both stable and reusable for unseen task combinations.
    
    \item We introduce GaTaP, which separates fast, task-specific routing through sparse gain inference from slower, shared representation learning through target propagation.
    
    \item We derive a closed-form, sparse update for the task-specific gains and a probabilistic target-inference rule that extends from multilayer perceptron to convolutional architectures.
    
    \item We show experimentally that GaTaP retains previously learned tasks while adapting to unseen task compositions with frozen shared weights, and that related tasks exhibit similar inferred gating patterns.
\end{itemize}

\section{Related Work}

\paragraph{Continual learning.}
Continual-learning methods are commonly grouped into regularization-, replay-, and architecture-based approaches. Regularization methods constrain updates to parameters considered important for previous tasks, as in Elastic Weight Consolidation (EWC)~\citep{kirkpatrick2017overcoming} and Synaptic Intelligence (SI)~\citep{zenke2017continual}. Replay-based approaches instead revisit stored or generated examples from previous tasks, including GEM~\citep{lopez2017gradient}, A-GEM~\citep{chaudhry2018efficient}, iCaRL~\citep{rebuffi2017icarl}, and Dark Experience Replay~\citep{buzzega2020dark}. While effective at mitigating catastrophic forgetting, these approaches primarily preserve previous knowledge rather than explicitly organizing it into computational components that can be recombined for novel tasks.

\paragraph{Modular networks and task-dependent routing.}
Architectural approaches reduce interference by assigning different parameters or computational pathways to different tasks. Progressive Neural Networks~\citep{rusu2016progressive} add and freeze task-specific columns, whereas PathNet~\citep{fernando2017pathnet}, PackNet~\citep{mallya2018packnet}, and Hard Attention to the Task~\citep{serra2018overcoming} select, reserve, or protect subsets of a larger network. In many such approaches, catastrophic forgetting is mitigated partly by construction: previously allocated parameters are frozen or future updates are restricted according to a predefined routing or capacity-allocation mechanism. While effective at reducing interference, this shifts an important part of the learning problem---determining which components are relevant to a new task and should therefore be reused or protected---into the architecture or training rules.

Particularly related to our work, RPS-Net~\citep{rajasegaran2020adaptiverandompathselection} progressively searches for task-specific paths through a modular network while encouraging the reuse of previously selected paths, thereby enabling forward transfer and reducing interference. Veniat et al.~\citep{veniat2021efficientcontinuallearningmodular} further propose Modular Networks with Task-Driven Priors (MNTDP), in which tasks are represented as compositions of reusable modules. A data-driven prior restricts architecture search to perturbations of paths associated with related previous tasks, enabling efficient transfer and improved scaling to long task sequences.

GaTaP shares the motivation of modular reuse but treats pathway selection itself as an inference problem. Rather than prescribing a discrete architecture or a fixed rule specifying which subnetworks should be protected or recruited, GaTaP represents task relevance through sparse multiplicative gains $\Gamma$ inferred directly from a support set. These gains determine which shared subnetworks contribute to the current task and, consequently, which components participate strongly in learning. Thus, the routing and selective-plasticity rules are encoded in learned task-dependent variables rather than imposed externally. At test time, the same mechanism allows new gain configurations to be inferred for previously unseen task compositions while all shared weights remain frozen.

\paragraph{Compositional generalization.}
Compositional generalization concerns the ability to systematically recombine previously learned components to solve novel task combinations~\citep{ZHOU2024105711, lake2017generalization, lake2015human}. Although modular continual-learning methods naturally promote parameter reuse, transfer to a new task does not necessarily imply compositional generalization~\citep{mittalModularArchitectureEnough2022,mittalDoesLearningRight2025}. We explicitly evaluate whether components acquired during sequential learning can be recombined without modifying the shared representation. For example, after learning to discriminate between MNIST~\citep{lecun2010mnist} digits $3$ and $4$, and between digits $5$ and $6$, GaTaP is evaluated on unseen combinations such as $3$ vs.\ $6$ by inferring only a new gating configuration. This provides a direct test of whether continual learning has produced reusable computational structure rather than merely preserved task-specific solutions.

\paragraph{Target propagation and local credit assignment.}
Target propagation provides an alternative to standard backpropagation by assigning desired activation states, or \emph{targets}, to intermediate layers~\citep{bengio2014auto,lee2015difference}. Difference Target Propagation~\citep{lee2015difference}, for instance, propagates corrected activation targets through learned approximate inverses rather than propagating only loss gradients. GaTaP builds on difference target propagation by deriving hidden targets through local Gaussian conditioning under a linearized network model, yielding Kalman-gain updates related to Gauss-Newton target methods~\citep{NEURIPS2020_e7a425c6}.

\section{Gated Target Propagation}
\label{sec:method}

GaTaP separates continual learning into two coupled processes operating at different timescales: \emph{fast task-specific routing} (inner loop), controlled by multiplicative gains $\Gamma$, and \emph{slow shared representation learning} (outer loop), stored in weights $W$. Given a task $\mathcal{T}$, a support set (training examples) is used to infer a sparse gating configuration $\Gamma_{\mathcal{T}}^\star$, while a disjoint query set is used to update the shared subnetworks with the inferred gates fixed. At inference time, $W$ is frozen and adaptation to a task occurs only through re-inference of $\Gamma$. Learning a task can thus happen in two ways: either through shaping shared weights $W$ during dual phase training, or through gain adaptation alone at inference. 

\subsection{Gated probabilistic network}
\label{sec:prob_model}

Consider an $L$-layer network with hidden states $h_i\in\mathbb{R}^{d_i}$, where $h_0=x$ denotes the input. Rather than using a single transformation at layer $i$, GaTaP contains $n_i$ shared subnetworks $\{W_i^{(k)}\}_{k=1}^{n_i}$. Their contributions are controlled by task-specific multiplicative gains $\Gamma_i^{(k)}\in\mathbb{R}^{d_i}$:
\begin{equation}
h_i =
\sum_{k=1}^{n_i}
\Gamma_i^{(k)}
\odot
\left(
W_i^{(k)}\phi_{i-1}(h_{i-1})
\right)
+\epsilon_i,
\qquad
\epsilon_i\sim\mathcal{N}(0,\Sigma_i),
\label{eq:gated_forward}
\end{equation}
where $\phi_{i-1}$ is the activation function and $\odot$ denotes elementwise multiplication. Thus, $\Gamma_i^{(k)}$ gates the output units of the $k$-th subnetwork. We define the corresponding task-dependent effective map as
\begin{equation}
W_i(\Gamma_i)
=
\sum_{k=1}^{n_i}
\operatorname{Diag}\!\left(\Gamma_i^{(k)}\right)W_i^{(k)},
\label{eq:effective_weight}
\end{equation}
so that
\begin{equation}
h_i
=
W_i(\Gamma_i)\phi_{i-1}(h_{i-1})+\epsilon_i.
\label{eq:compact_forward}
\end{equation}
The subnetworks $W_i^{(k)}$ are shared across tasks, whereas the gains $\Gamma_i^{(k)}$ are inferred separately for each task. A gain close to zero suppresses the contribution of the corresponding unit or subnetwork, thereby reducing its participation both in the forward computation and in subsequent learning.

We interpret~\Eqref{eq:compact_forward} probabilistically as
\begin{equation}
p(h_i\mid h_{i-1},\Gamma_i,W_i)
=
\mathcal{N}
\left(
W_i(\Gamma_i)\phi_{i-1}(h_{i-1}),
\Sigma_i
\right).
\label{eq:conditional_layer}
\end{equation}
Under this model, hidden states and gains are latent variables that can be inferred using both the input and the desired output. This provides the probabilistic basis for target propagation: rather than communicating only a gradient from the output loss, we infer corrected hidden states $\{h_i^\star\}_{i=1}^{L}$ and use them as local learning targets. A Laplace prior on the gains, $p(\Gamma_i)\propto \exp( \alpha_i\|\Gamma_i\|_1)$, encourages sparse task-dependent routing. Full derivations of the corresponding variational objective are provided in Appendix~\ref{subsubsec:linearonehiddenlayer}.

\begin{figure}[t]
    \centering
    \includegraphics[width=12cm]{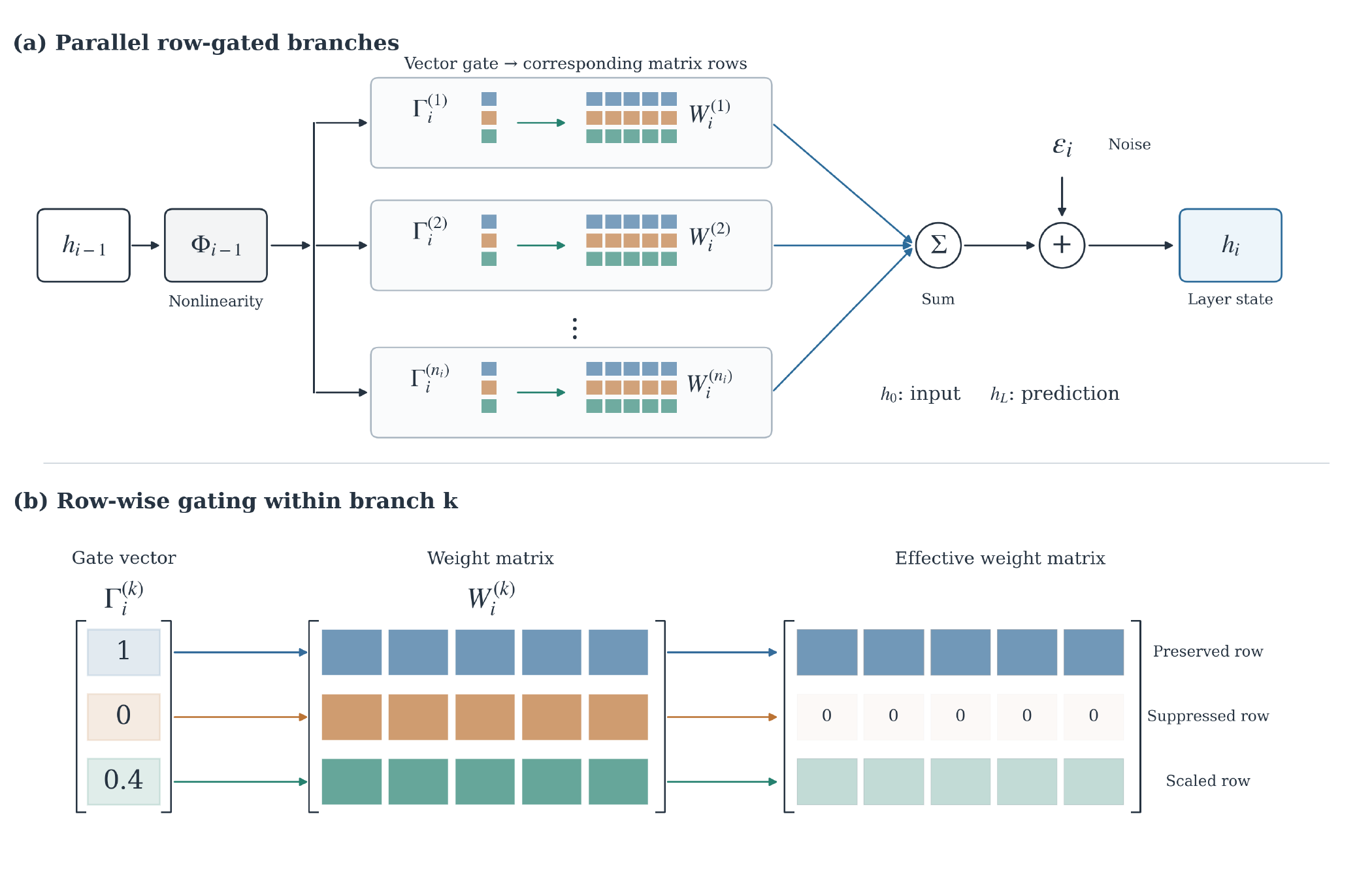}
    \caption{
    \textbf{Parallel subnetworks with task-specific row-wise gating.}
    (a) The transformed state $\Phi_{i-1}(h_{i-1})$ is processed
    by $n_i$ parallel subnetworks with shared weight matrices
    $W_i^{(k)}$. Each task-specific gate vector $\Gamma_i^{(k)}$
    scales the rows of its corresponding weight matrix.
    The resulting branch outputs are summed, and additive noise
    $\epsilon_i$ yields the next layer state $h_i$.
    (b) Illustration of row-wise gating within one subnetwork:
    each gate entry multiplies every weight in the corresponding
    row. Values of $1$, $0$, and $0.4$ respectively preserve,
    suppress, and attenuate that row, allowing task-dependent
    modulation of shared connectivity.
    }
    \label{fig:rowwise-gating}
\end{figure}

\subsection{Probabilistic target inference}
\label{sec:target_inference}

Target propagation requires a desired state $h_i^\star$ for each intermediate layer. GaTaP obtains these targets through approximate probabilistic inference. Consider the forward mapping from layer $i$ to $i+1$,
\begin{equation}
F_{i+1}(h_i)
=
W_{i+1}(\Gamma_{i+1})\phi_i(h_i),
\qquad
h_{i+1}
=
F_{i+1}(h_i)+\epsilon_{i+1}.
\label{eq:forward_map}
\end{equation}
For a linear-Gaussian mapping, the conditional posterior over $h_i$ given a desired state at the layer above is Gaussian, and its posterior mean can be computed analytically using a Kalman update. For nonlinear networks, we locally linearize $F_{i+1}$ around the current feedforward state $h_i$. Let
\begin{equation}
J_{i+1}
=
\left.
\frac{\partial F_{i+1}(h)}
     {\partial h}
\right|_{h=h_i}
\label{eq:jacobian}
\end{equation}
denote the local Jacobian. The resulting extended-Kalman target update is:
\begin{equation}
h_i^\star
=
h_i+
K_i
\left(
h_{i+1}^\star-h_{i+1}
\right),
\label{eq:target_update}
\end{equation}
with
\begin{equation}
K_i
=
\Sigma_iJ_{i+1}^{\top}
\left(
J_{i+1}\Sigma_iJ_{i+1}^{\top}
+
\Sigma_{i+1}
\right)^{-1}.
\label{eq:kalman_gain}
\end{equation}
The output target is initialized from the prediction error,
\begin{equation}
h_L^\star
=
h_L+K_L(y-h_L),
\label{eq:output_target}
\end{equation}
and~\Eqref{eq:target_update} is then applied recursively from the output toward earlier layers.

\Eqref{eq:target_update} has a simple interpretation: the inferred target is the current feedforward state plus a correction determined by the discrepancy at the next layer. The gain matrix $K_i$ controls how strongly this discrepancy should modify $h_i$ according to the uncertainty assigned to adjacent layers. Unlike standard backpropagation, the signal delivered to layer $i$ is therefore a target state $h_i^\star$ rather than only the gradient of a global loss. The linear-Gaussian derivation along with its nonlinear extension are given in Appendices~\ref{subsec:derivationkalmanupdate} and ~\ref{subsec:nonlinearneuralnet}.

\subsection{Fast, sparse gate inference}
\label{sec:gamma_inference}

Given the target states, GaTaP infers which shared subnetworks should be recruited for the current training epoch, without explicit knowledge of which task is being performed. Importantly, this adaptation modifies only $\Gamma$, while the shared weights remain fixed.

Let
$\mathcal{S}_{\mathcal{T}}
=\{(x_p,y_p)\}_{p=1}^{P}$
be the support set for a given training epoch. For layer $i$, define the response of subnetwork $k$ to support example $p$ as
\begin{equation}
Z_{i,p}^{(k)}
=
W_i^{(k)}
\phi_{i-1}(h_{i-1,p}).
\label{eq:subnetwork_response}
\end{equation}
The gated prediction of the target state is:
\begin{equation}
h_{i,p}
=
\sum_{k=1}^{n_i}
\Gamma_i^{(k)}
\odot Z_{i,p}^{(k)}.
\end{equation}
We infer the gains by fitting the locally inferred targets while imposing an
$\ell_1$ penalty:
\begin{equation}
\mathcal{L}_{i,\mathrm{support}}(\Gamma_i)
=
\frac{1}{2P}
\sum_{p=1}^{P}
\left\|
h_{i,p}^\star
-
\sum_{k=1}^{n_i}
\Gamma_i^{(k)}
\odot Z_{i,p}^{(k)} - \epsilon_{i}
\right\|_2^2
+
\alpha_i
\sum_{k=1}^{n_i}
\|\Gamma_i^{(k)}\|_1.
\label{eq:gamma_objective}
\end{equation}
The sparsity penalty corresponds to a Laplace prior over $\Gamma$ and favors task-specific solutions that recruit only a subset of the available components.

Although~\Eqref{eq:gamma_objective} couples the gains of multiple subnetworks, coordinate descent reduces each update to a scalar $\ell_1$ regularized least-squares problem. For neuron $q$ of subnetwork $k$, define the partial residual:
\begin{equation}
R_{i,p,q}^{(-k)}
=
h_{i,p,q}^{\star}
-
\sum_{\ell\neq k}
\gamma_{i,q}^{(\ell)}
Z_{i,p,q}^{(\ell)}- \epsilon_{i},
\label{eq:partial_residual}
\end{equation}
where $\gamma_{i,q}^{(\ell)} \in \Gamma_{i}^{(\ell)}$ is the gain variable specific to neuron $q$ for subnetwork $\ell$. Further, define the sufficient statistics:
\begin{equation}
a_{i,k,q}
=
\frac{1}{P}
\sum_{p=1}^{P}
\left(Z_{i,p,q}^{(k)}\right)^2,
\qquad
b_{i,k,q}
=
\frac{1}{P}
\sum_{p=1}^{P}
Z_{i,p,q}^{(k)}
R_{i,p,q}^{(-k)}.
\label{eq:gamma_stats}
\end{equation}
Given these, the coordinate-wise optimum has the closed form:
\begin{equation}
\boxed{
\left(\gamma_{i,q}^{(k)}\right)^\star
=
\operatorname{sign}(b_{i,k,q}) \times \frac{
\left\lfloor \left\lvert b_{i,k,q}\right\rvert - \alpha_i \right\rfloor^{+}
}{
a_{i,k,q}
},}
\label{eq:gamma_closed_form}
\end{equation}
where $\left\lfloor z \right\rfloor^{+}=\max(z,0)$. Thus, components whose support-set evidence does not exceed the sparsity threshold are assigned exactly zero gain. The resulting $\Gamma^\star$ therefore defines a sparse computational route specific to the selected task for that training epoch, while blocking learning signal flow backward through inactive pathways during learning. The complete derivation can be found in Appendix~\ref{CloseformsolutionFFNs}.

\subsection{Shared-weight learning}
\label{sec:bilevel}

After inferring $\Gamma^\star$ from the support set, the gains are held fixed and the shared weights are optimized on a disjoint query set $\mathcal{Q}$. For each query example, target states are first computed using Equations \ref{eq:target_update}-\ref{eq:kalman_gain}. The local objective at layer $i$ is
\begin{equation}
\mathcal{L}_{i,\mathrm{query}}
=
\frac{1}{2|\mathcal{Q}_{\mathcal{T}}|}
\sum_{(x,y)\in\mathcal{Q}_{\mathcal{T}}}
\left\|
h_i^\star
-
\sum_{k=1}^{n_i}
\left(\Gamma_i^{(k)}\right)^\star
\odot
W_i^{(k)}
\phi_{i-1}(h_{i-1}) - \epsilon_{i}
\right\|_{\Sigma_i^{-1}}^2 ,
\label{eq:query_loss}
\end{equation}
where $\|v\|_{\Sigma^{-1}}^2=v^\top\Sigma^{-1}v$. Only the shared parameters are updated:
\begin{equation}
W_i^{(k)}
\leftarrow
W_i^{(k)}
-
\eta
\nabla_{W_i^{(k)}}
\mathcal{L}_{i,\mathrm{query}}.
\label{eq:weight_update}
\end{equation}
Training can therefore be summarized as the bi-level procedure:
\begin{equation}
\Gamma^\star(W)
=
\arg\min_{\Gamma}
\left[
\mathcal{L}_{\mathrm{support}}
(\Gamma;W,\mathcal{T})
+
\sum_{i,k}
\alpha_i\|\Gamma_i^{(k)}\|_1
\right],
\label{eq:inner}
\end{equation}
followed by
\begin{equation}
W
\leftarrow
W
-
\eta
\nabla_W
\mathcal{L}_{\mathrm{query}}
\left(
W,
\Gamma_{\mathcal{T}}^\star(W);
\mathcal{T}
\right).
\label{eq:outer}
\end{equation}
The two stages play distinct roles: the inner problem rapidly determines which shared components should be recruited for the current task, while the outer problem updates the associated shared weights. Because the same shared parameters are reused across tasks, this decomposition allows task-specific computation without maintaining an independent network for every task. Refer to Algorithm~\ref{alg:gated_targetprop} for the full training procedure.

\paragraph{Inference on seen and unseen tasks.}
After the complete continual-learning sequence, the shared parameters are frozen:
\begin{equation}
W \leftarrow W^\star,
\qquad
\Delta W = 0.
\end{equation}
For an evaluation task $\mathcal{T}$, either previously seen or an unseen composition, only $\Gamma^\star$ is computed according to~\Eqref{eq:inner}. Performance is then measured on a disjoint query set. Consequently, performance on an unseen task cannot be attributed to further learning of the shared representation. Instead, the model rapidly reconfigures previously acquired components through few-shot inference of $\Gamma^\star$.

\subsection{Convolutional extension}
\label{sec:cnn_extension}

To scale GaTaP beyond fully connected networks, we extend the same gating and target-propagation mechanism to convolutional architectures by replacing each linear subnetwork response with a convolution. For feature map $h_{i-1}$,
\begin{equation}
h_i = R_i \left( h_{i-1} \right) + \sum_{k=1}^{n_i} \Gamma_i^{(k)} \odot \left( W_i^{(k)} * \phi_{i-1}(h_{i-1}) \right) +\epsilon_i,
\label{eq:cnn_gating}
\end{equation}
where $*$ denotes the convolution operation and $R_i$ is the residual connection. Additionally, gains can be applied channel-wise, using one gain per output channel shared across all spatial locations, or spatially, using location-specific gains that modulate different positions within each feature map. The sparse coordinate update in~\Eqref{eq:gamma_closed_form} remains unchanged in form, with its sufficient statistics additionally aggregated over spatial positions for channel-wise gating. The target update also retains the form of~\Eqref{eq:target_update}. In convolutional networks, the Kalman gain becomes computationally expensive due to the large matrix inversion in $(J_i\Sigma_iJ_i^\top+\Sigma_{i+1})^{-1}$. We therefore use a diagonal approximation in our convolutional experiments that renders inversion trivial. Further details are provided in Appendix~\ref{subsec:ConvolutionalNeuralNetworks}.

\section{Experimental Setup}
\label{sec:experiments}
Our experiments are designed to test the two central claims of the proposed framework: whether GaTaP can retain knowledge acquired sequentially, and whether the resulting shared components can be recombined to solve novel task compositions. We therefore focus on controlled class-incremental classification experiments in which both retention and compositional reuse can be measured directly.

\paragraph{Datasets and architectures.}
We evaluate GaTaP on MNIST, CIFAR-10, and CIFAR-100, progressively increasing the visual complexity and number of classes. MNIST experiments use a feedforward gated network, whereas CIFAR-10 and CIFAR-100 use convolutional architectures. The main experiments are formulated as sequences of binary classification tasks. Additional preliminary multi-class experiments are reported in Supplemental Figure~\ref{fig:retention_across_tasks}; these results are intended as an initial test of scalability beyond the binary setting, and a more systematic multi-class evaluation is left for future work. Architecture and hyperparameter details are provided in Appendix~\ref{Hyperparameters}.

\paragraph{Class-incremental task construction.}
All main experiments use the class-incremental learning setting: tasks are presented sequentially, and the original class identities are preserved throughout learning. For example, an MNIST training stream may sequentially require discrimination between digits 1 and 2, 3 and 4, 5 and 6, and 7 and 8. The model has access only to the current task data during its corresponding training stage. We distinguish between \emph{seen tasks}, which are presented during sequential training, and \emph{unseen compositional tasks}, which recombine classes encountered in different training tasks. For instance, after learning $1$ vs.\ $2$ and $3$ vs.\ $4$, tasks such as $1$ vs.\ $4$ or $2$ vs.\ $3$ require a new combination of previously acquired components.

\paragraph{Training and evaluation protocol.}
For each task $\mathcal{T}_j$, its data are divided into a support set $\mathcal{S}_j$ and a disjoint query set $\mathcal{Q}_j$. During training, $\mathcal{S}_j$ is used to infer the task-specific gains $\Gamma_j^\star$, while $\mathcal{Q}_j$ is used to update the shared weights $W$ through target propagation. After the complete task sequence, all shared weights are frozen. Evaluation on both seen and unseen tasks is then performed by inferring only a new gain configuration from the task support set and measuring performance on its query set. Consequently, performance on unseen compositions cannot be attributed to further learning of the shared representation.

\paragraph{Evaluation criteria.}
Retention is measured by the final accuracy on previously seen tasks.
Compositional generalization is measured by query accuracy on unseen class combinations with the shared weights fixed. We additionally analyze the cosine similarity between inferred gain vectors to determine whether tasks sharing classes recruit similar computational routes, and vary the support-set size to evaluate few-shot inference of gain variables. GaTaP-trained models were compared to equivalent architectures trained with backpropagation and EWC (with code adapted from \citep{chua2026balancing}).

\section{Results}
\label{sec:results}

\paragraph{Retention and compositional generalization.}

\begin{figure}[t]
    \centering
    \includegraphics[width=0.48\textwidth]{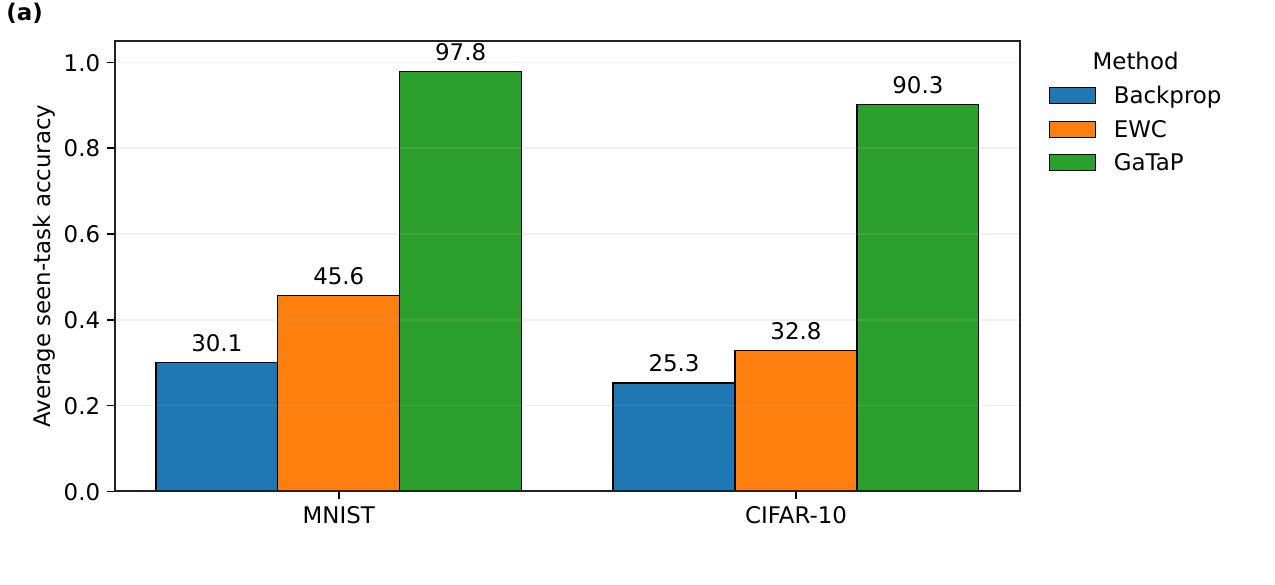}
    \hfill
    \includegraphics[width=0.48\textwidth]{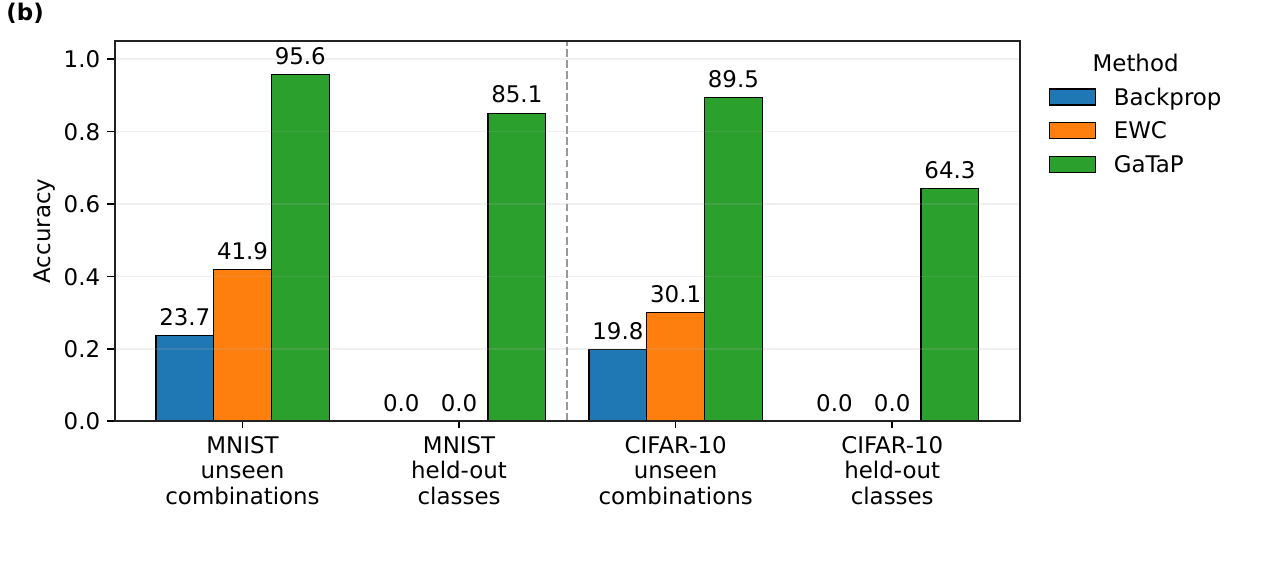}

    \vspace{4pt}

    \includegraphics[height=6cm,width=0.48\textwidth]{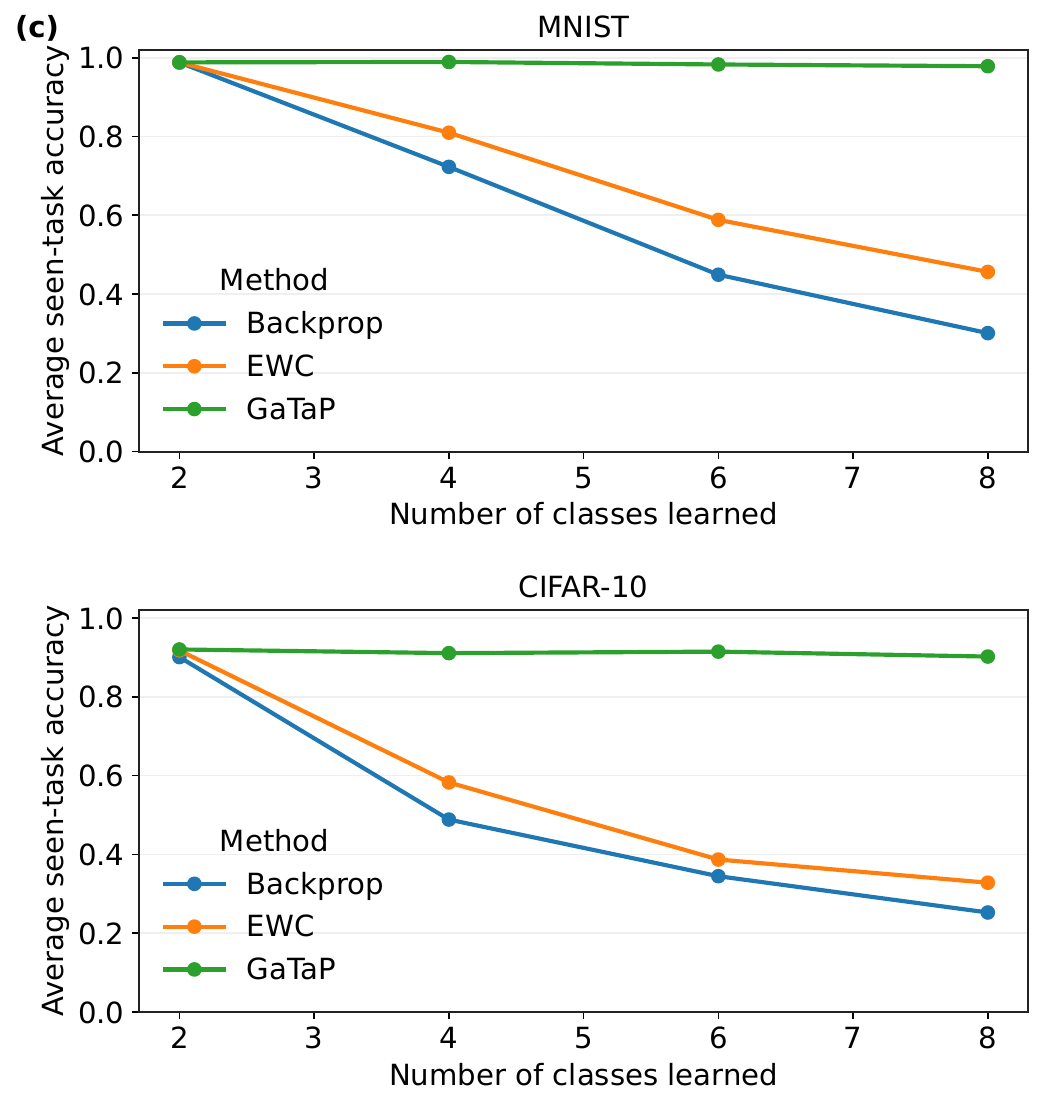}
    \hfill
    \includegraphics[height=6cm,width=0.48\textwidth]{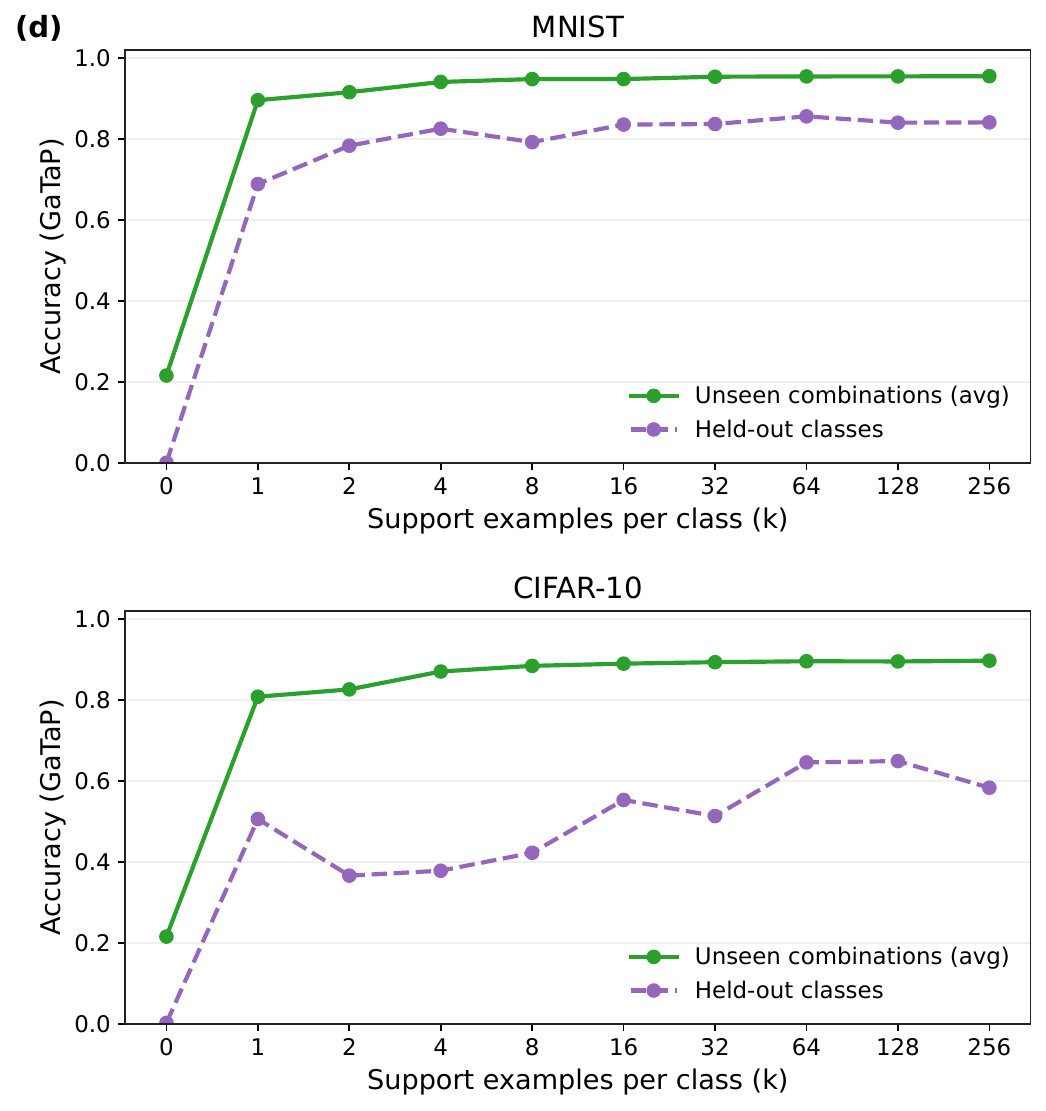}

    \caption{MNIST vs.\ CIFAR-10 comparison across methods. \textbf{a)} Average performance on tasks networks have already been exposed to during the training process, evaluated at the end of training. \textbf{b)} Average performance on unseen task combinations of data classes already seen during training, compared to performance on task combinations never seen during training. \textbf{c)} Same as (a), evaluated as networks are trained on an increasing number of tasks for MNIST (top) and CIFAR-10 (bottom). \textbf{d)} Average performance on unseen task combinations as a function of the number of support set examples provided per-class to train gates $\Gamma$ during evaluation, for MNIST (top) and CIFAR-10 (bottom).}
    \label{fig:mnist_cifar10_comparison}
\end{figure}

 After training, all shared weights $W$ were frozen; only the task-specific gains $\Gamma_{\mathcal{T}}$ were inferred from a support set, and performance was measured on a disjoint query set. On seen tasks (Fig.~\ref{fig:mnist_cifar10_comparison}a), GaTaP reached $0.978$ final average accuracy on MNIST, compared to $0.301$ (Backprop) and $0.456$ (EWC); similar gaps held on CIFAR-10. This indicates that GaTaP effectively protects learned representations compared to alternative methods. The same frozen representation also generalized to unseen task combinations (Fig.~\ref{fig:mnist_cifar10_comparison}.b): with accuracy reaching $0.956$ on unseen digit combinations and $0.851$ on a held-out task never seen during training (MNIST), and $0.895$ / $0.643$ on CIFAR-10. Since $W$ was never updated at evaluation time, this adaptation came solely from inferring a new gain configuration, via a single closed-form optimization step, using 64 support examples; Fig.~\ref{fig:mnist_cifar10_comparison}d shows comparable accuracy with far fewer examples shown. This supports the central claim that the learned components can be recombined for few-shot generalization through a single, close-form optimization step.

\begin{figure}[t]
    \centering
    \includegraphics[width=0.48\textwidth]{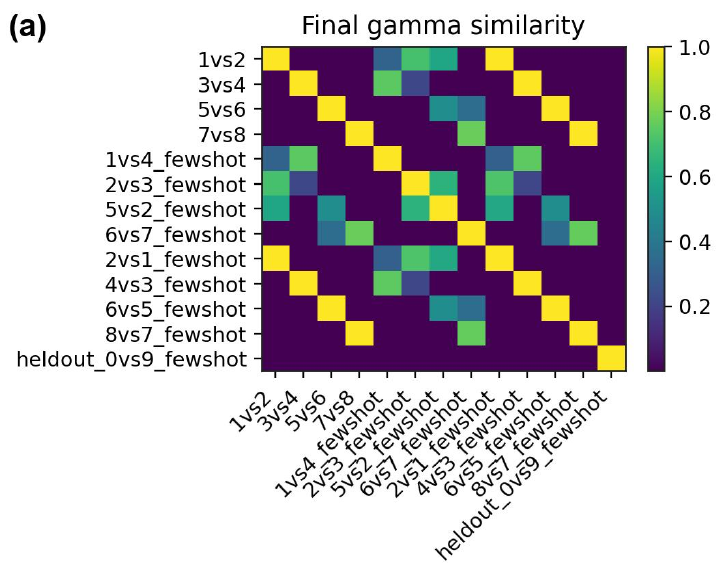}
    \hfill
    \includegraphics[width=0.48\textwidth]{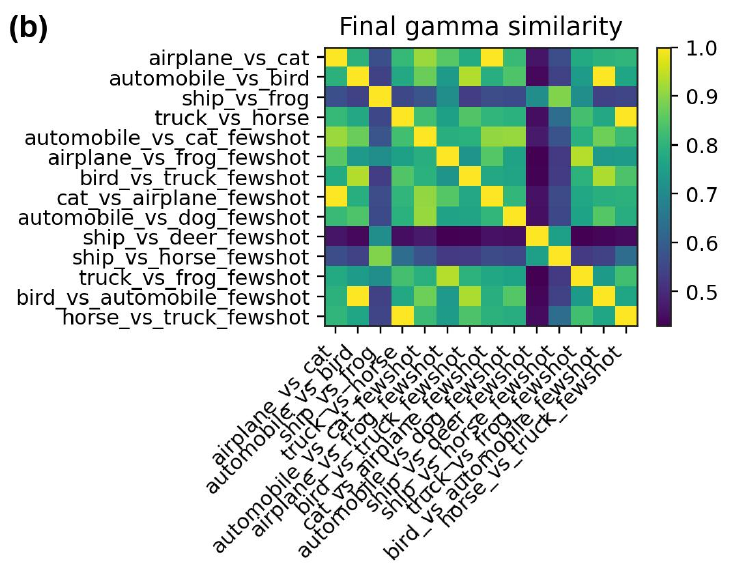}

    \vspace{4pt}

    \includegraphics[height=4.5cm,width=0.48\textwidth]{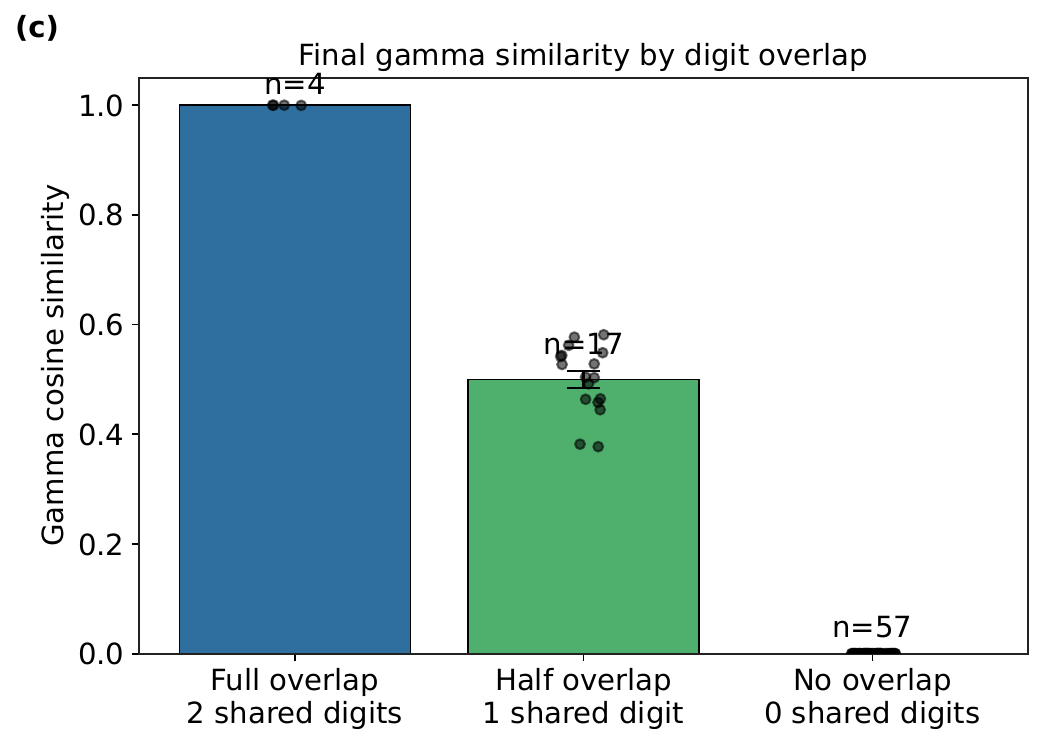}
    \hfill
    \includegraphics[height=4.5cm,width=0.48\textwidth]{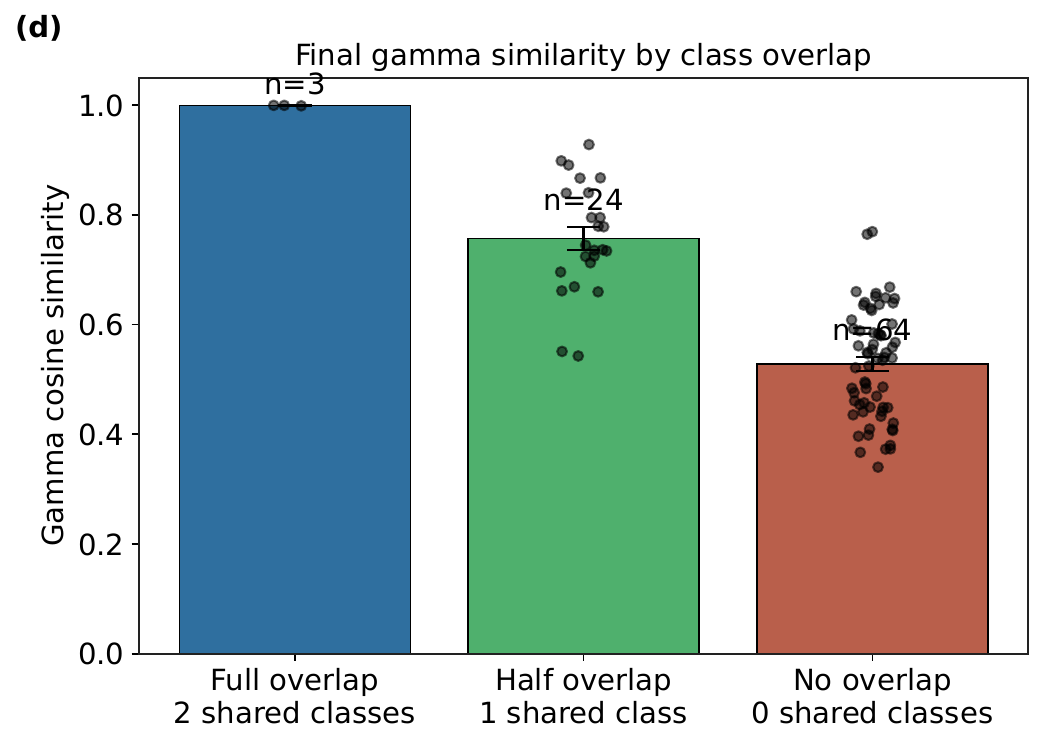}

    \caption{Visualizing gain similarity across tasks. \textbf{a)} Cosine similarity of learned $\Gamma$ parameters across MNIST two-way classification tasks. ``fewshot" indicates an unseen recombination of previously seen digits. \textbf{b)} Same as (a), for CIFAR-10 two-way classification tasks. \textbf{c)} Average cosine similarity of $\Gamma$ parameters for MNIST two-way classification tasks that share both digits, one digit, or no digits. \textbf{d)} Same as (c), for CIFAR-10 two-way classification tasks.}
    \label{fig:gamma_similarity}
\end{figure}

\paragraph{Structure of the inferred gating configurations.}
We next investigated whether the inferred task-specific gains encoded meaningful relationships between tasks. If the gains identified reusable computational components, then tasks sharing classes should be expected to recruit more similar routes than unrelated tasks. We therefore computed the cosine similarity between the final gain configurations of different task pairs and grouped them according to their degree of class overlap.
%We then investigate if inferred gains encode task relationships, by computing cosine similarity between gain configurations grouped by class overlap. 
As shown in Fig.~\ref{fig:gamma_similarity}, $\Gamma$ similarity tracked task overlap closely on MNIST (near-identical for matching tasks, near-zero for disjoint ones), and a similar trend was observed on CIFAR-10. This indicates that related tasks reuse related computational pathways, while unrelated tasks recruit more distinct ones.

\section{Discussion and Limitations} 
\label{sec:discussion}
In this study we have introduced \textit{gated target propagation} (GaTaP), a learning algorithm capable of both preventing catastrophic forgeting and enabling compositional generalization through a shared mechanism: the rapid optimization of network gains $\Gamma$ across epochs. Our method does not require explicit knowledge of task transitions---only output errors---and it outperforms both standard back propagation and EWC on continual learning benchmarks, while simultaneously supporting few-shot compositional generalization to task recombinations and wholly unseen classification tasks.

Our method can be seen as a generalization of difference target propagation \citep{lee2015difference}, as well as other methods that use Kalman-style error corrections to provide closed-loop modification of internal network states prior to learning. Previous work, generally restricted to \textit{additive}---as opposed to our \textit{multiplicative}---corrections, have shown that this form of closed-loop control can provide benefits for learning stability, particularly in chaotic dynamical systems \citep{saugtekin2026error, kaleb2024feedback, sussillo2009generating}. Our work builds on these methods by showing how similar error-based feedback signals can be used in concert with gain modulation mechanisms to enable continual learning and compositional generalization. Collectively, these studies demonstrate that error signals used for both control and learning can provide important improvements relative to standard backpropagation-based methods.

Interestingly, methods closely related to GaTaP, such as difference target propagation and control-theoretic learning methods, have been proposed as models of learning and synaptic plasticity in the brain \citep{podlaski2020biological, gilra2017predicting, alemi2018learning}. While our method relies on biologically implausible operations to compute feedback errors (e.g. calculating a Kalman gain matrix), previous work has demonstrated that feedback projections can still enable learning when they adapt through local plasticity to dynamically approximate the analytically optimal feedback signal \citep{lee2015difference,meulemans2021credit}. Therefore, relating GaTaP more concretely to models of learning in the brain remains a promising direction for future work.

The present study nevertheless has several limitations. First,
computing the Kalman-style target update involves Jacobian--covariance operations that
become expensive in larger networks. Our convolutional network implementation uses
a diagonal approximation (Appendix \ref{subsec:ConvolutionalNeuralNetworks}) to make this computation tractable, introducing a
trade-off between inference fidelity and computational efficiency. Methods to approximate or amortize the computation of feedback errors may greatly increase the computational efficiency of our method. Finally,
the current experiments focus on static visual tasks. Extending the framework to temporal models, larger architectures, and more complex task compositions will be important for assessing its generality.

\section{Conclusion}
\label{sec:conclusion}
We have introduced GaTaP, a continual-learning framework that jointly addresses knowledge retention and compositional reuse. The method separates fast, task-specific adaptation from slower, shared learning: sparse, multiplicative gains infer which subnetworks are relevant to a task, while target propagation updates the shared weights through local target states. After continual training, the shared weights remain fixed and both previously learned and unseen task compositions are solved through gain inference alone, a simple closed form update. Experiments on MNIST, CIFAR-10, and CIFAR-100 show that GaTaP retains previously acquired knowledge, recombines learned components for novel tasks, and produces gating patterns that reflect task structure. These results suggest that continual learning and compositional generalization can be jointly enabled through a shared learning architecture based on rapidly reconfiguring multiplicative gains.

\subsection*{Acknowledgements}
AMN and CB would like to thank the members of the GL and BAR's labs. 
% GL and BAR acknowledge support from the Canada-CIFAR AI Research Chair program. GL acknowledges further support from the Canada Research Chair in Neural Computations and Interfacing (CIHR Tier 2) and NSERC Discovery grant RGPIN-2026-05516
This work was supported by: [GL] NSERC Discovery Grant (RGPIN-2026-05516), Canada CIFAR AI Chair Program, Canada Research Chair in Neural Computations and Interfacing (CIHR, tier 2). [BAR] NSERC (Discovery Grant: RGPIN-2020-05105; Discovery Accelerator Supplement: RGPAS-2020-00031; Arthur B. McDonald Fellowship: 566355-2022) and CIFAR (Canada AI Chair; Learning in Machine and Brains Fellowship). The authors acknowledge the material support of NVIDIA in the form of computational resources.

\subsection*{AI use statement}
In this work, we used generative AI tools to generate images of our architecture.

\bibliography{iclr2027_conference}
\bibliographystyle{iclr2027_conference}

\newpage

\appendix
\renewcommand{\thefigure}{S\arabic{figure}} % This makes figures numbered as S1, S2, etc.

\setcounter{figure}{0} % Reset figure counter

\section{Appendix}
\label{Appendix}
% You may include other additional sections here.
\subsection{One Hidden Layer Probabilistic Model}
\label{subsec:linearonehiddenlayer}
In this section, we describe the proposed gated target propagation framework starting from a single layer neural network model, before progressing to deeper and convolutional models. The main idea is to separate fast task adaptation from slow weight learning within a unified and theoretically justifiable framework.

We first consider a neural network with a single hidden layer. Let \(s\in \mathbb{R}^{d_0}\) denote the input stimulus, \(r\in \mathbb{R}^{d_1}\) the hidden state, and \(o\in \mathbb{R}^{d_2}\) the output. The input and output weights are denoted by \(W_1\) and \(W_2\), respectively. We also include task-dependent gain variables \(\Gamma_1\) and \(\Gamma_2\). These variables multiplicatively modulate the activity of the network. In our model, the gain parameters act as a gating mechanism: they select which hidden units or subnetworks are active for a given task. This will allow the network to reuse shared parameters while adapting rapidly to different tasks.

For a single subnetwork per layer, the stochastic neuron equations are
\begin{align}
    r = \Gamma_1 \odot \left(W_1 s\right) + \epsilon_1,
    \qquad
    \epsilon_1 \sim \mathcal{N}(0,\sigma_1^2 I),
\end{align}
and
\begin{align}
    o = \Gamma_2 \odot \left(W_2 r\right) + \epsilon_2,
    \qquad
    \epsilon_2 \sim \mathcal{N}(0,\sigma_2^2 I).
\end{align}
where $\odot$ is the Hadamard (element-wise) product. Equivalently, we can define a conditional model probability distribution $P_{M}$ that defines joint distribution of hidden states $r$ and outputs $o$ conditioned on $\Gamma$ and $W$ parameters:
\begin{align}
    P_M(r \mid s, \Gamma, W)
\sim
\mathcal{N}
\left(
\Gamma_1 \odot \left(W_1 s\right),
\sigma_1^2 I
\right),
\end{align}
and
\begin{align}
    P_M(o \mid r, s, \Gamma, W)
\sim
\mathcal{N}
\left(
\Gamma_2 \odot \left(W_2 r\right),
\sigma_2^2 I
\right).
\end{align}
We refer to $W_1$ and $W_2$ as subnetworks or connection weights between two consecutive layers. In a more general subnetwork formulation, two consecutive layers are connected by multiple subnetworks. For instance, $W_i^{(k)}$ with $k=1,\ldots,n_i$, are the connections weights between layer $i-1$ and layer $i$. Note that in this framework there are considerably fewer gain parameters than there are weight parameters: for the hidden layer, $\Gamma_1 \in \mathbb{R}^{d_1}$, whereas $W_1 \in \mathbb{R}^{d_2} \times \mathbb{R}^{d_1}$. We next define the effective weights as follows:
\begin{align}
    \overline{W}_1 \left(\Gamma_1 \right)
=
\sum_{k=1}^{n_1}
\text{diag}\left( \Gamma_1^{(k)}  \right) W_1^{(k)},
\qquad
\overline{W}_2 \left(\Gamma_2 \right)
=
\sum_{k=1}^{n_2}
\text{diag}\left( \Gamma_2^{(k)}  \right) W_2^{(k)}.
\end{align}
The neuron equations can then be written as
\begin{align} \label{eq:r_effective}
    r
=
\overline{W}_1 \left(\Gamma_1 \right) s
+
\epsilon_1,
\end{align}
and
\begin{align} \label{eq:o_effective}
    o
=
\overline{W}_2 \left(\Gamma_2 \right) r
+
\epsilon_2.
\end{align}
Note that if the gain for a particular unit $r_j$ is zero ($\Gamma_{1,j} = 0$), then the effective weights for that unit $\overline{W}_{1,j,:}$ will also be zero. We will be adapting gain parameters rapidly on a per-task basis, while the weights will adapt on a slower timescale. By encouraging sparsity in our gain parameters for each task, we will in turn `protect' weight parameters from parameter updates on tasks where they are less relevant for task performance.

\subsubsection{Objective function}
\label{subsubsec:linearonehiddenlayer}

Let's consider a shallow neural network, where $\overline{W}_1\left( \Gamma_1\right)$ and $\overline{W}_2\left( \Gamma_2\right)$ are the effective weights:
\begin{align}
    \overline{W}_1\left( \Gamma_1\right) = \sum_{k=1}^{n_1}  \text{diag}\left( \Gamma_1^{(k)}  \right) W_{1}^{(k)} \text{ and } \overline{W}_2\left( \Gamma_2\right) = \sum_{k=1}^{n_2}  \text{diag}\left( \Gamma_2^{(k)}  \right) W_{2}^{(k)}.
\end{align}
Let $s \in \mathbb{R}^{d_0}$ be the input stimuli, $r$ the hidden state, and $o$ the output, governed by \Eqref{eq:r_effective} and \Eqref{eq:o_effective}, respectively. For $\Gamma = (\Gamma_1, \Gamma_2)$, we assume $P_M(\Gamma)$ (marginalizing across tasks) is Laplace-distributed, which promotes sparsity for individual tasks. We further assume, for each task $\mathcal{T}$, that the data pairs \((s,o)\) are generated from a joint data distribution \(P_{\mathcal{D}}(s,o | \mathcal{T})\), which in practice correspond to input and classification target pairs. Our goal is to learn a model distribution \(P_M(o|s)\) that matches the conditional data distribution \(P_{\mathcal{D}}(o\mid s, \mathcal{T})\), where here we suppress dependence of $P_M$ on the network parameters $W = (W_1, W_2)$ for conciseness and assume that the model does not have direct access to the task $\mathcal{T}$ that is being performed. Ideally, we would want to minimize the objective function:
\begin{equation}
\mathcal{J} = \mathbb{E}_{s,\mathcal{T}} \left [D_{KL} [P_{\mathcal{D}}(o|s, \mathcal{T}) || P_M(o|s)] \right ],
\end{equation}
which would simply correspond to matching network inputs and outputs to labelled pairs, averaged across tasks and stimuli sampled from the data distribution. However, at this point, for our neural network we have only defined the conditional distribution $P_M(o,r|s, \Gamma)$, whereas the loss is defined in terms of the marginal distribution over $r$ and $\Gamma$:
\begin{align}
\label{initial_objective}
    P_M(o|s) &= \iint P_M(o,r|s,\Gamma) P_M(\Gamma) dr d\Gamma.
\end{align}
Computing this probability requires marginalizing over the latent gain variables \(\Gamma\) and the hidden state $r$ to make \(P_M(o|s)\). Since this marginalization is in practice analytically intractable, we will instead resort to minimizing a well-known upper bound on our loss, known as the \textit{variational free energy} \citep{bernardo2003variational}, which will prove much easier to compute. 

To do this, we first introduce a simplified variational distribution \(Q(r,\Gamma|s,o)\) to approximate the true posterior \(P_M(r,\Gamma|s,o)\). Essentially, this posterior provides a correction to the network latent states when both stimuli \textit{and} outputs have been observed. We use our variational distribution to define the variational free energy as:  
\begin{equation}
\mathcal{L} = \mathbb{E}_{s,\mathcal{T}} \biggl [D_{KL}\Bigl [P_{\mathcal{D}}(o|s, \mathcal{T}) Q(r, \Gamma|s, o) || P_M(o, r, \Gamma|s) \Bigr ] \biggr].
\end{equation}
To see that $\mathcal{L}$ is an upper bound for $\mathcal{J}$, observe:
\begin{align}
    \mathcal{L} &= \mathbb{E}_{s,\mathcal{T}} \biggl [D_{KL} \Bigl [P_{\mathcal{D}}(o|s, \mathcal{T}) Q(r, \Gamma|s, o) || P_M(o, r, \Gamma|s) \Bigr ] \biggr]\\
    % &= \iiint P_{\mathcal{D}}(o|s) Q(r, \Gamma|s, o) \log \left( \frac{P_{\mathcal{D}}(o|s) Q(r, \Gamma|s, o)}{P_M(o, r, \Gamma|s)}\right) dr do d\Gamma\\
    % &= \iiint P_{\mathcal{D}}(o|s) Q(r, \Gamma|s, o) \log \left( \frac{P_{\mathcal{D}}(o|s) Q(r, \Gamma|s, o)}{P_M(o|s) P_M( r, \Gamma|s, o)}\right) dr do d\Gamma\\
    % &= \iiint \left( P_{\mathcal{D}}(o|s) Q(r, \Gamma|s, o) \log \left( \frac{P_{\mathcal{D}}(o|s) }{P_M(o|s)}\right) + P_{\mathcal{D}}(o|s) Q(r, \Gamma|s, o) \log \left( \frac{Q(r, \Gamma|s, o)}{P_M( r, \Gamma|s, o)}\right) \right) dr do d\Gamma\\
    % &= \iint Q(r, \Gamma|s, o)dr d\Gamma \times \int P_{\mathcal{D}}(o|s) \log \left( \frac{P_{\mathcal{D}}(o|s) }{P_M(o|s)}\right) do  + \iiint \left( P_{\mathcal{D}}(o|s) Q(r, \Gamma|s, o) \log \left( \frac{Q(r, \Gamma|s, o)}{P_M( r, \Gamma|s, o)}\right) \right) dr do d\Gamma\\
    % &= \int P_{\mathcal{D}}(o|s) \log \left( \frac{P_{\mathcal{D}}(o|s) }{P_M(o|s)}\right) do + \iiint \left( P_{\mathcal{D}}(o|s) Q(r, \Gamma|s, o) \log \left( \frac{Q(r, \Gamma|s, o)}{P_M( r, \Gamma|s, o)}\right) \right) dr do d\Gamma\\
    &= \mathbb{E}_{s,\mathcal{T}} \biggl [ D_{KL} \Bigl [P_{\mathcal{D}}(o|s, \mathcal{T}) || P_M(o|s) \Bigr ] \biggr ] + \mathbb{E}_{o,s,\mathcal{T}} \biggl [ D_{KL} \Bigl [Q(r, \Gamma|s, o) || P_M(r, \Gamma|s, o) \Bigr] \biggr] \\
    &\geq \mathcal{J}.
\end{align}
Where the inequality follows from the the positivity of the Kullback-Leibler divergence, i.e. the fact that $\mathbb{E}_{o,s,\mathcal{T}} \left( D_{KL} [Q(r, \Gamma|s, o) || P_M(r, \Gamma|s, o)] \right) \geq 0$. To understand this upper bound, note that we are now effectively performing two optimizations: approximating the posterior over network states $(r,\Gamma)$, and subsequently approximating the input-output probability distribution. In the event that $Q$ corresponds to the true posterior, the bound is tight, i.e. $\mathcal{J} = \mathcal{L}$. This suggests a coordinate descent procedure called \textit{variational expectation maximization}, where one first minimizes $\mathcal{L}$ with respect to $Q$ (restricted to some convenient class of probability distributions), and then one subsequently minimizes $\mathcal{L}$ with respect to network parameters $W$. To make this process more concrete, we will first simplify our loss function $\mathcal{L}$:
\begin{align}
    \mathcal{L} &= \mathbb{E}_{s,\mathcal{T}} \biggl [D_{KL}\Bigl [P_{\mathcal{D}}(o|s, \mathcal{T}) Q(r, \Gamma|s, o) || P_M(o, r, \Gamma|s) \Bigr ] \biggr]\\
    &= \mathbb{E}_{s,\mathcal{T}} \left [\iiint P_{\mathcal{D}}(o|s,\mathcal{T}) Q(r, \Gamma|s, o) \log \left( \frac{P_{\mathcal{D}}(o|s) Q(r, \Gamma|s, o)}{P_M(o, r, \Gamma|s)}\right) dr do d\Gamma \right ]\\
    &\equiv \mathbb{E}_{s,o,\mathcal{T}, Q} \Bigl [\log  Q(r, \Gamma|s, o) - \log  P_M(o, r, \Gamma|s) \Bigr ]\\
    &= \mathbb{E}_{s,o,\mathcal{T}, Q} \Bigl [- \log  P_M(o, r, \Gamma|s) \Bigr ]+ \mathbb{E}_{s,o,\mathcal{T}, Q} \Bigl [H ( Q(r, \Gamma|s, o)) \Bigr ],
\end{align}
where we use $\mathbb{E}_Q$ to indicate expectations over $r$ and $\Gamma$ variables sampled from the approximate posterior, and $H ( Q(r, \Gamma|s, o))$ is the entropy of $Q$. We will choose a maximum \textit{a posteriori} (MAP) point estimate for our $Q$ distribution over $r$ and $\Gamma$, where we assume $\Gamma$ is an arbitrary constant for each task $\mathcal{T}$ and $r$ is allowed to vary with stimuli $s$. Because point estimates have constant entropy, we can ignore the entropy term for the purposes of optimization, i.e.:
\begin{equation}\label{eq:nll_loss}
    \mathcal{L} \equiv \mathbb{E}_{s,o,\mathcal{T}} \Bigl [- \log  P_M \left ( o, r^*, \Gamma^*|s \right ) \Bigr],
\end{equation}
where $r^*$ is the hidden state estimate for stimulus $s$, and $\Gamma^*$ is the gain estimate for all stimuli sampled from the support set for a given training epoch; we describe how these estimates are calculated in more detail below. The resultant loss is simply the negative log-likelihood of $P_M$, evaluated at inferred hidden network states $r^*$ and $\Gamma^*$. This leads to the following expanded loss, substituting in the conditional $\log$-probability distributions for the hidden and output network layers, along with the sparse Laplace prior term for $\Gamma$ gating variables:
\begin{align}
    \mathcal{L} &= \mathbb{E}_{s,o,\mathcal{T}} [- \log P_M(r^*\mid s, \Gamma^*_1) - \log P_M(o\mid r^*, \Gamma^*_2) - \log P(\Gamma^*_1, \Gamma^*_2)]\\
    &= \mathbb{E}_{s,o,\mathcal{T}} \Bigg[ \frac{1}{2\sigma_1^2}
\left\|
r^*
-
\overline W_1(\Gamma^*_{1})s
\right\|_2^2
+
\frac{1}{2\sigma_2^2}
\left\|
o-
\overline W_2(\Gamma^*_{2})
r^*
\right\|_2^2
+
\alpha_1
\|\Gamma^*_{1}\|_1
+
\alpha_2
\|\Gamma^*_{2}\|_1
\Bigg],
\end{align}
where $\sigma^2_1$, $\sigma^2_2$, $\alpha_1$, and $\alpha_2$ are model hyperparameters.
This loss has to be minimized for $r^*$, $\Gamma^*$, and $W$. There are many ways to carry out this optimization procedure in principle; in what follows, we will detail one particular method with convenient properties, based on coordinate descent. We observe that for network states $r^*$ this optimization reduces to a simple, local regression problem if $W$ and $\Gamma^*$ are held fixed---the same holds true for $W$ and $\Gamma^*$. Exploiting this fact, we employ the following coordinate descent scheme (detailed in Algorithm \ref{alg:gated_targetprop}). For each epoch of training, network states $r^*$ and multiplicative gain parameters $\Gamma$ are inferred from a small support set. Subsequently, we hold $\Gamma^*$ fixed, and optimize for network states $r^*$ and parameters $W$ on the remainder of the data. Interestingly, this procedure does not require explicit knowledge of whether the task $\mathcal{T}$ has changed for the given epoch: optimal $\Gamma^*$ variables simply change when dictated by changes in the statistics of $(s,o)$ pairs, relying only on the assumption that the task remains the same for the duration of each epoch.

Below, we explain exactly how optimization occurs for $r^*,\Gamma^*,$ and $W$. Weight updates are closely related to difference target propagation ~\citep{bengio2014auto, lee2015difference}, where layer is assigned a local target state $r^*$ and trained toward it with a local, per-layer loss instead of a single backpropagated global gradient; we adapt this by deriving each layer's target via Kalman-like update rather than the autoencoder-based inverse mappings of classical difference target propagation~\citep{lee2015difference}. At test time, the shared weights are frozen and only the gamma parameters are inferred again, to assess the model's ability to generalize to new data.

Unlike standard backpropagation-based learning, GaTaP gives the model two distinct adaptation mechanisms. The gamma parameters provide task-specific routing, while the shared subnetworks store reusable knowledge. Therefore, if two tasks share useful structure, they may activate similar gamma patterns even if they are not identical tasks.

\subsubsection{Derivation of the target using the Kalman update in the linear case}
\label{subsec:derivationkalmanupdate}
In this section, we provide the methodology for inferring the MAP estimate $r^*$, conditioned on $W, \Gamma$ and the input-output pair $(s,o)$. While methods based on predictive coding \citep{rao1999predictive, whittington2017approximation} would perform gradient descent on \Eqref{eq:nll_loss} to calculate $r^*$, we will instead analytically compute the posterior distribution for $r^*$ as well as its mode. In the linear Gaussian case, computing this posterior is simple; in the \textit{non}linear Gaussian case, we can approximate the posterior via a Taylor expansion, as detailed in the next section (Appendix \ref{subsec:nonlinearneuralnet}).

\textbf{Recall: Conditional Gaussian Distribution}\\\\
For a joint Gaussian distribution,
\begin{align}
    \begin{bmatrix}x\\y\end{bmatrix} \sim \mathcal{N} \left( \begin{bmatrix}\mu_x\\\mu_y\end{bmatrix}, \begin{bmatrix}
\Sigma_{xx} & \Sigma_{xy} \\\Sigma_{yx} & \Sigma_{yy} \end{bmatrix}\right),
\end{align}
the conditional distribution is given by:
\begin{equation} \label{eq:gaussian_cond}
x|y \sim \mathcal{N}\left(\mu_x + \Sigma_{xy}\Sigma_{yy}^{-1}(y-\mu_y),  \Sigma_{xx} - \Sigma_{xy}\Sigma_{yy}^{-1}\Sigma_{yx}\right).\end{equation}

\textbf{Kalman update}

As in the previous section, we assume the following:
\begin{align}
    P_M(r\mid s, \Gamma_1)
=
\mathcal{N}
\left(
\overline{W}_1 \left(\Gamma_1 \right) s,
\Sigma_1
\right) \qquad \text{ and } \qquad  P_M(o\mid r, \Gamma_2)
=
\mathcal{N}
\left(
\overline{W}_2 \left(\Gamma_2 \right)r,
\Sigma_2
\right),
\end{align}
Where $\Sigma_1 = \epsilon_1 I$ and $\Sigma_2 = \epsilon_2 I$. Thus, the model defines the joint distribution
\begin{align}
    P_M(r,o\mid s, \Gamma)
=
P_M(o\mid r, \Gamma_2)P_M(r\mid s, \Gamma_1).
\end{align}
Because both distributions are Gaussian and the mapping from \(r\) to \(o\) is linear, the conditional joint distribution of \(r\) and \(o\) given \(s\) is also Gaussian:
\begin{align}
    P_M
\left(
\begin{bmatrix}
r\\
o
\end{bmatrix}
\middle| s, \Gamma
\right)
=
\mathcal{N}
\left(
\begin{bmatrix}
\overline{W}_1 \left(\Gamma_1 \right) s\\
\overline{W}_2 \left(\Gamma_2 \right) \overline{W}_1 \left(\Gamma_1 \right) s
\end{bmatrix},
\begin{bmatrix}
\Sigma_1
&
\Sigma_1 \overline{W}_2 \left(\Gamma_2 \right)^T
\\
\overline{W}_2 \left(\Gamma_2 \right)\Sigma_1
&
\overline{W}_2 \left(\Gamma_2 \right)\Sigma_1 \overline{W}_2 \left(\Gamma_2 \right)^T+\Sigma_2
\end{bmatrix}
\right).
\end{align}
Using this joint distribution and \Eqref{eq:gaussian_cond}, we can infer the mean of the hidden state conditioned on the observed stimulus and output as well as the fixed $\Gamma$ variables:
\begin{align}
    \mathbb{E}_{p_M(r\mid o,s,\Gamma)}[r]
=
\overline{W}_1 \left(\Gamma_1 \right) s
+
\Sigma_1 \overline{W}_2 \left(\Gamma_2 \right)^T
\left(
\overline{W}_2 \left(\Gamma_2 \right)\Sigma_1 \overline{W}_2 \left(\Gamma_2 \right)^T+\Sigma_2
\right)^{-1}
\left(
o-\overline{W}_2 \left(\Gamma_2 \right)\overline{W}_1 \left(\Gamma_1 \right) s
\right).
\end{align}
We define the Kalman gain
\begin{align}
    K
=
\Sigma_1 \overline{W}_2 \left(\Gamma_2 \right)^T
\left(
\overline{W}_2 \left(\Gamma_2 \right)\Sigma_1 \overline{W}_2 \left(\Gamma_2 \right)^T+\Sigma_2
\right)^{-1}.
\end{align}
Then the posterior mean can be written more compactly as
\begin{align}
    \mathbb{E}_{p_M(r\mid o,s)}[r]
=
\overline{W}_1 \left(\Gamma_1 \right) s
+
K
\left(
o-\hat{o}
\right),
\end{align}
where
\begin{align}
    \hat{o}
=
\overline{W}_2 \left(\Gamma_2 \right)\overline{W}_1 \left(\Gamma_1 \right) s
\end{align}
is the feedforward prediction of the output. For a Gaussian posterior, the mean and the mode are the same. Therefore, the posterior mean gives the optimal inferred hidden state under the linear Gaussian model. We denote this inferred state by
\begin{align}
    r^{\star}
=
\bar r
+
K
\left(
o-\hat{o}
\right),
\end{align}
where $\bar r = \overline{W}_1 \left(\Gamma_1 \right) s$.
This expression shows that the inferred hidden target is the mean unconditioned feedforward hidden state plus an additive correction term. The correction depends on the output error
\begin{align}
    e_o = o-\hat{o},
\end{align}
and the Kalman gain \(K\) determines how much of this output error should be assigned to the hidden state.

This derivation provides the basic target propagation rule in the linear case. The weights are optimized to minimize the mean squared error between the the actual hidden state provided by the model and the target. The target hidden state is not obtained by backpropagating gradients through the network. Instead, it is inferred probabilistically as the hidden state that best `explains' both the input and the observed output under the model.

\subsection{Nonlinear one hidden layer neural network: Extended Kalman update}
\label{subsec:nonlinearneuralnet}
The previous derivation relied on a linear relationship between the hidden state and the output. In this section, we show how the same Kalman-style target update can be extended to a nonlinear one-hidden-layer network by using a local linear approximation, similar to the extended Kalman filter~\citep{ribeiro2004kalman}.

We consider the model
\begin{align}
    r \mid s &\sim \mathcal{N}\left(\overline{W}_1 \left(\Gamma_1 \right) s, \Sigma_1\right), \\
    o \mid r &\sim \mathcal{N}\left(\overline{W}_2\left(\Gamma_2 \right) f(r), \Sigma_2\right),
\end{align}
where $s$ is the input, $r$ is the hidden state, $o$ is the output, $f$ is a nonlinear activation function, and $\overline{W}_2$ maps the activated hidden state to the output.

Because $f(r)$ is nonlinear, the posterior distribution $p_M(r \mid o,s)$ is not Gaussian in general. Therefore, we approximate the nonlinear mapping locally. Let $r_0$ be the current feedforward hidden state, and let
\begin{align}
    J_f = \left.\frac{\partial f}{\partial r}\right|_{r=r_0}
\end{align}
be the Jacobian of the activation function evaluated at $r_0$. Using a first-order Taylor expansion,
\begin{align}
    f(r) \approx f(r_0) + J_f(r-r_0).
\end{align}
Substituting this approximation into the output model gives
\begin{align}
    o \approx \overline{W}_2\left(\Gamma_2 \right) f(r_0) + \overline{W}_2\left(\Gamma_2 \right) J_f(r-r_0) + \epsilon_2,
    \qquad
    \epsilon_2 \sim \mathcal{N}(0,\Sigma_2).
\end{align}
For simplicity, define the local linear map
\begin{align}
    H = \overline{W}_2\left(\Gamma_2 \right) J_f.
\end{align}
Then the linearized output model becomes
\begin{align}
    o \approx \overline{W}_2\left(\Gamma_2 \right) f(r_0) + H(r-r_0) + \epsilon_2.
\end{align}
Under this local linear approximation, the joint distribution of $r$ and $o$ conditioned on $s$ is approximately Gaussian:
\begin{align}
    p_M\left(
    \begin{bmatrix}
        r \\
        o
    \end{bmatrix}
    \middle| s
    \right)
    \approx
    \mathcal{N}
    \left(
    \begin{bmatrix}
        r_0 \\
        \overline{W}_2\left(\Gamma_2 \right) f(r_0)
    \end{bmatrix},
    \begin{bmatrix}
        \Sigma_1 & \Sigma_1 H^T \\
        H\Sigma_1 & H\Sigma_1 H^T + \Sigma_2
    \end{bmatrix}
    \right).
\end{align}
Therefore, the mean of the approximate posterior distribution $p_M(r \mid o,s)$ is
\begin{align}
    \mathbb{E}_{p_M(r \mid o,s)}[r]
    \approx
    r_0
    +
    \Sigma_1 H^T
    \left(
        H\Sigma_1 H^T + \Sigma_2
    \right)^{-1}
    \left(
        o - \overline{W}_2\left(\Gamma_2 \right) f(r_0)
    \right).
\end{align}
Replacing $H$ by $\overline{W}_2\left(\Gamma_2 \right)J_f$, we obtain the nonlinear target update
\begin{align}
    r^\star
    =
    r_0
    +
    \Sigma_1 J_f^T \overline{W}_2\left(\Gamma_2 \right)^T
    \left(
        \overline{W}_2\left(\Gamma_2 \right) J_f \Sigma_1 J_f^T \overline{W}_2\left(\Gamma_2 \right)^T + \Sigma_2
    \right)^{-1}
    \left(
        o - \overline{W}_2\left(\Gamma_2 \right) f(r_0)
    \right).
\end{align}
Equivalently, defining the Kalman gain
\begin{align}
    K
    =
    \Sigma_1 J_f^T \overline{W}_2\left(\Gamma_2 \right)^T
    \left(
        \overline{W}_2\left(\Gamma_2 \right) J_f \Sigma_1 J_f^T \overline{W}_2\left(\Gamma_2 \right)^T + \Sigma_2
    \right)^{-1},
\end{align}
the hidden target can be written as
\begin{align}
    r^\star = r_0 + K\left(o-\hat{o}\right),
    \qquad
    \hat{o} = \overline{W}_2\left(\Gamma_2 \right) f(r_0).
\end{align}
Thus, as in the linear case, the inferred target is the current feedforward hidden state plus an error-correction term. The main difference is that the linear weight matrix $\overline{W}_2$ is replaced by the local linearization $\overline{W}_2 J_f$. This gives an analytic approximation to the optimal hidden state, instead of requiring iterative gradient descent to infer the target. Specifically, this single-step closed-form update mirrors the Gauss-Newton formulation of target propagation~\citep{NEURIPS2020_e7a425c6}, where the regularized pseudo-inverse of the local Jacobian is used to analytically jump to the target rather than iteratively optimizing toward it.

\subsection{Target propagation in deep networks with multiplicative gains}
\label{subsec:DeepLearningFrameworks}
The previous section showed how to infer a target state for a nonlinear network with one hidden layer. We now extend the same idea to a deep network, where a target state must be inferred at every layer. Let $h_0 \in \mathbb{R}^{d_0}$ denote the input and let $h_i \in \mathbb{R}^{d_i}$ denote the hidden state at layer $i$. For a feedforward network with $L$ layers, we define the gated hidden state as
\begin{equation}
    h_i
    =
    \sum_{k=1}^{n_i}
    \Gamma_i^{(k)} \odot \left( W_i^{(k)} \phi_{i-1}(h_{i-1}) \right) + \epsilon_{i}, \qquad\text{ with } \epsilon_{i} \sim \mathcal{N} \left( 0, \Sigma_{i}\right),  \qquad  i = 1,\ldots,L.
\end{equation}
Here, $n_i$ is the number of subnetworks connecting layer $i-1$ to layer $i$, $W_i^{(k)} \in \mathbb{R}^{d_i \times d_{i-1}}$ is the $k$-th subnetwork weight matrix, $\phi_{i-1}$ is the activation function applied to the previous hidden state and $\phi_{0}$ is just the identity function. The vector $\Gamma_i^{(k)} \in \mathbb{R}^{d_i}$ gates the rows of $W_i^{(k)}$. Equivalently, we can define an effective task-dependent weight matrix
\begin{equation}
    \overline{W}_i
    =
    \sum_{k=1}^{n_i}
    \Gamma_i^{(k)} \odot W_i^{(k)},
\end{equation}
so that
\begin{equation}
    h_i
    =
    \overline{W}_i \phi_{i-1}(h_{i-1}) 
    + \epsilon_{i}, \qquad\text{ with } \epsilon_{i} \sim \mathcal{N} \left( 0, \Sigma_{i}\right),  \qquad  i = 1,\ldots,L.
\end{equation}
The weights $W_i^{(k)}$ are shared across tasks, while the gain parameters are inferred separately in each training epoch (agnostic to whether or not the task $\mathcal{T}$ has changed). In the next sections, we describe how to compute MAP estimates for the gains $\Gamma^*$ and the network hidden states $h_i^*$.

\subsubsection{Close-form solution for the $\Gamma_{i}^{(k)}$.}
\label{CloseformsolutionFFNs}

The closed-form update for $\Gamma$ enables fast and sparse task adaptation without an additional gradient-based inner optimization. Inferring $\Gamma^\star$ before updating the shared weights first determines the task-specific computational route, after which the outer optimization modifies the corresponding shared components while leaving less relevant pathways unaffected. This separation supports fast task inference while reducing interference during slower shared-weight learning.

Let consider the equation: $h_{i} = \sum_{k=1}^{n_i} \Gamma_{i}^{(k)}  \odot \left( W_{i}^{(k)}  \phi_{i-1}\left(h_{i-1}\right) \right) + \epsilon_{i},\,\,i = 2, \ldots, L.\,$ Here, $\Gamma_{i}^{(k)} = \left[\gamma_{i,1}^{(k)},\ldots, \gamma_{i,d_i}^{(k)} \right], h_i \in \mathbb{R}^{d_i}$ and $W_{i}^{(k)} \in \mathbb{R}^{d_i\times d_{i-1}}.$ Let $a_{i-1,p} =  \phi_{i-1}\left(h_{i-1,p}\right)$ for a data point $p$ in a batch of $P$ data points. For subnetwork $k$, define $Z_{i,p}^{(k)} = W_{i}^{(k)}  \phi_{i-1}\left(h_{i-1,p}\right)$ so that $h_{i,p} = \sum_{k=1}^{n_i} \Gamma_{i}^{(k)} \odot Z_{i,p}^{(k)}$. Let assume for layer $i$ we have the target $h_{i,p}^{*}$ (we will discuss how these are obtained in the next section, using \Eqref{targetupdate}). The local loss (on the support set) is defined as follows:
\begin{align}
    \mathcal{L}_{i, \text{support}} \left( \Gamma_{i}^{(k)}\right) &= \dfrac{1}{2P} \sum_{p=1}^{P} \left\lVert h_{i,p}^{*} - h_{i,p}\right\rVert_{2}^{2} + \alpha_{i} \sum_{k=1}^{n_i} \left\lVert \Gamma_{i}^{(k)} \right\rVert_{1}\\
    &= \dfrac{1}{2P} \sum_{p=1}^{P} \left\lVert h_{i,p}^{*} - \sum_{k=1}^{n_i} \Gamma_{i}^{(k)} \odot Z_{i,p}^{(k)} - \epsilon_{i}\right\rVert_{2}^{2} + \alpha_{i} \sum_{k=1}^{n_i} \left\lVert \Gamma_{i}^{(k)} \right\rVert_{1}\\
    &= \dfrac{1}{2P} \sum_{p=1}^{P} \sum_{q=1}^{d_i} \left( h_{i,p,q}^{*} - \sum_{k=1}^{n_i} \gamma_{i,q}^{(k)} Z_{i,p,q}^{(k)} - \epsilon_{i}\right)^2 + \alpha_{i} \sum_{k=1}^{n_i} \sum_{q=1}^{d_i} \left\lvert \gamma_{i,q}^{(k)} \right\rvert.
\end{align}
For a fixed output hidden neuron $q$, we have: 
\begin{align}
    \mathcal{L}_{i,k,q} \left( \gamma_{i,q}^{(k)}\right) &= \dfrac{1}{2P} \sum_{p=1}^{P} \left( R_{i,p,q}^{-k} - \gamma_{i,q}^{(k)} Z_{i,p,q}^{(k)}\right)^2 + \alpha_{i} \left\lvert \gamma_{i,q}^{(k)} \right\rvert + \sum_{l\neq k}^{n_i} \left\lvert \gamma_{i,q}^{(l)} \right\rvert,
\end{align}
with
\begin{align}
    \mathcal{L}_{i} \left( \Gamma_{i}^{(k)}\right) = \sum_{q=1}^{d_i} \mathcal{L}_{i,k,q} \left( \gamma_{i}^{(k)}\right) \text{ and } R_{i,p}^{-k} = h_{i,p}^{*} - \sum_{l\neq k}^{d_i}\Gamma_{i}^{(l)} \odot Z_{i,p}^{(l)} - \epsilon_{i}.
\end{align}
The optimum of $\mathcal{L}_{i,k,q} \left( \gamma_{i,q}^{(k)}\right)$ should either be zero, or $\gamma_{i,q}^{(k)}$ should be zero. We first assume that $\gamma_{i,q}^{(k)}$ is nonzero. Then the derivative of $\mathcal{L}_{i,k,q} \left( \gamma_{i,q}^{(k)}\right)$ with respect to $\gamma_{i,q}^{(k)}$ is given by
\begin{align}
    \nabla_{\gamma_{i,q}^{(k)}} \mathcal{L}_{i,k,q} \left( \gamma_{i,q}^{(k)}\right) &= -\dfrac{1}{P}\sum_{p=1}^{P} Z_{i,p,q}^{(k)} \left( R_{i,p,q}^{-k} - \gamma_{i,q}^{(k)} Z_{i,p,q}^{(k)}\right) + \alpha_{i}\times \text{sign} \left(\gamma_{i,q}^{(k)} \right) = 0\\
    &\implies \gamma_{i,q}^{(k)} = \dfrac{\dfrac{1}{P}\sum_{p=1}^{P} Z_{i,p,q}^{(k)} R_{i,p,q}^{-k} - \alpha_{i}\times \text{sign} \left(\gamma_{i,q}^{(k)} \right)}{\dfrac{1}{P}\sum_{p=1}^{P} \left( Z_{i,p,q}^{(k)}\right)^2}.
\end{align}
From this equality, we can infer the following three possible conditions:
\begin{align}
    & \text{If }\,\,\, \gamma_{i,q}^{(k)} > 0,  \dfrac{1}{P}\sum_{p=1}^{P} Z_{i,p,q}^{(k)} R_{i,p,q}^{-k} - \alpha_{i} > 0 \implies \dfrac{1}{P}\sum_{p=1}^{P} Z_{i,p,q}^{(k)} R_{i,p,q}^{-k} > \alpha_{i},\\
    & \text{else if} \,\,\, \gamma_{i,q}^{(k)} < 0, \dfrac{1}{P}\sum_{p=1}^{P} Z_{i,p,q}^{(k)} R_{i,p,q}^{-k} + \alpha_{i} < 0 \implies \dfrac{1}{P}\sum_{p=1}^{P} Z_{i,p,q}^{(k)} R_{i,p,q}^{-k} < -\alpha_{i} \\
    & \text{else} \,\,\, \gamma_{i,q}^{(k)} = 0.
\end{align}
Therefore, the close-form solution for $\gamma_{i,q}^{(k)}$ is given by:
\begin{align} \label{eq:closed_form_gamma}
    \left( \gamma_{i,q}^{(k)}\right)^{*} = \text{sign}\left( b_{i,k,q} \right)\times \dfrac{\left\lfloor \left\lvert b_{i,k,q}\right\rvert - \alpha_i \right\rfloor^{+}}{a_{i,k,q}}
\end{align}
where
\begin{align}
    a_{i,k,q} = \dfrac{1}{P}\sum_{p=1}^{P} \left( Z_{i,p,q}^{(k)}\right)^2 \,\,\text{ and }\,\, b_{i,k,q} = \dfrac{1}{P}\sum_{p=1}^{P} Z_{i,p,q}^{(k)} R_{i,p,q}^{-k}.
\end{align}
Therefore
\begin{align}
    \left( \Gamma_{i}^{(k)}\right)^{*} = \left[ \left( \gamma_{i,1}^{(k)}\right)^{*}, \ldots, \left( \gamma_{i,d_{i}}^{(k)}\right)^{*}\right].
\end{align}

In principle, calculating optimal $h_i^*$ and $\left (\Gamma_i^{(k)} \right )^*$ values should occur via a coordinate descent procedure, alternating repeatedly between computing the target states and gain variables across layers. In practice, however, we found that only a single iteration was needed for good performance, and so we took this approach to generate our empirical results for the sake of computational efficiency.

\subsubsection{Outer-loop optimization}
\label{outerloop_optim}
After gamma has been inferred from the support set, the shared weights are updated on the query set. The output target is defined as
\begin{equation}
    h_L^\star = h_L + K_L \left(y - h_L\right),
\end{equation}
where $y$ is the target label representation and $K_L$ controls the size of the output correction.

For hidden layers, the target is propagated backward through the network. Let
\begin{equation}
    F_{i+1}(h_i) = \overline{W}_{i+1} \phi_i(h_i)
\end{equation}
be the forward map from layer $i$ to layer $i+1$. Its Jacobian is
\begin{equation}
    J_i = \frac{\partial F_{i+1}(h_i)}{\partial h_i}.
\end{equation}
The hidden target is computed as
\begin{equation}
\label{targetupdate}
    h_i^\star = h_i + K_i \left(h_{i+1}^\star - h_{i+1}\right),
    \qquad i = L-1,\ldots,1,
\end{equation}
where
\begin{equation}
    K_i = \Sigma_i J_i^T \left(J_i \Sigma_i J_i^T + \Sigma_{i+1} \right)^{-1}.
\end{equation}
Here, $\Sigma_i = \sigma_i^2 I$ controls the uncertainty allowed at layer $i$.

Once the targets $h_i^\star$ and gains $\Gamma^\star$ are fixed, the update of $W_i$ depends only on the local query loss at layer i. The support objective, sparsity penalty on $\Gamma$, and losses from other layers are constant with respect to $W_i$, since gradients are not propagated through the inferred targets or gains. Hence, for the weight update,
\begin{align}
    \nabla_{W_i}\mathcal{L}_{\mathrm{total}} = \nabla_{W_i}\mathcal{L}_{i,\mathrm{query}},
\end{align}
up to any explicit regularization applied directly to $W_i$. Given these targets, the local weight-learning objective at layer $i$ is
\begin{equation}
    \mathcal{L}_{i, \text{query}} = \frac{1}{2P^{\prime}} \sum_{p=1}^{P^{\prime}} \left\| h_{i,p}^\star - \sum_{k=1}^{n_i} \left( \Gamma_i^{(k)}\right)^{\star} W_i^{(k)} \phi_{i-1}(h_{i-1,p}) - \epsilon_{i} \right\|_{\Sigma_i^{-1}}^2,
\end{equation}
where $P^{\prime}$ is the number of query examples. During this update, the gamma parameters are kept fixed. Only the shared weights $W_i^{(k)}$ are optimized.

The training procedure can therefore be summarized as
\begin{equation}
    \Gamma_T^\star
    =
    \arg\min_{\Gamma}
    \mathcal{L}_{\mathrm{support}}
    \left(
    \Gamma ; W, T
    \right)
    +
    \sum_{i,k}
    \alpha_i
    \left\|
    \gamma_i^{(k)}
    \right\|_1,
\end{equation}
followed by
\begin{equation}
    W
    \leftarrow
    W
    -
    \eta
    \nabla_W
    \mathcal{L}_{\mathrm{query}}
    \left(
    W, \Gamma_T^\star ; T
    \right).
\end{equation}
This is the modified inner-loop and outer-loop optimization used in the proposed framework. The support set is used only for gamma inference, and the query set is used only for updating the shared weights. 

At test time, the update of $W$ is removed. The weights are frozen, gamma is inferred from the support set, and accuracy and loss are computed on the query set. The full algorithm for the training procedure is given in [\ref{alg:gated_targetprop}].

\subsection{Convolutional Neural Networks}
\label{subsec:ConvolutionalNeuralNetworks}
\begin{figure}[p]
    \centering
    \includegraphics[width=\linewidth]{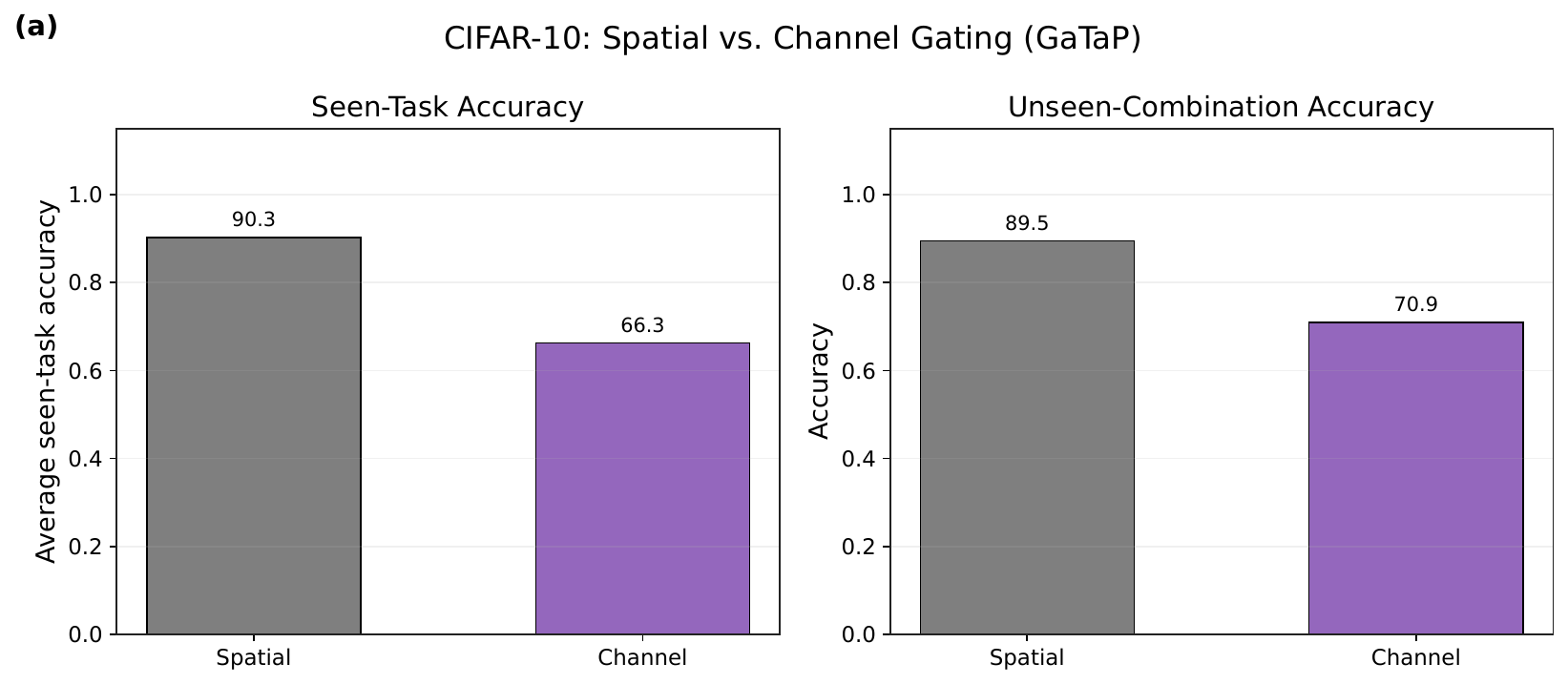}\\[4pt]
    \includegraphics[width=\linewidth]{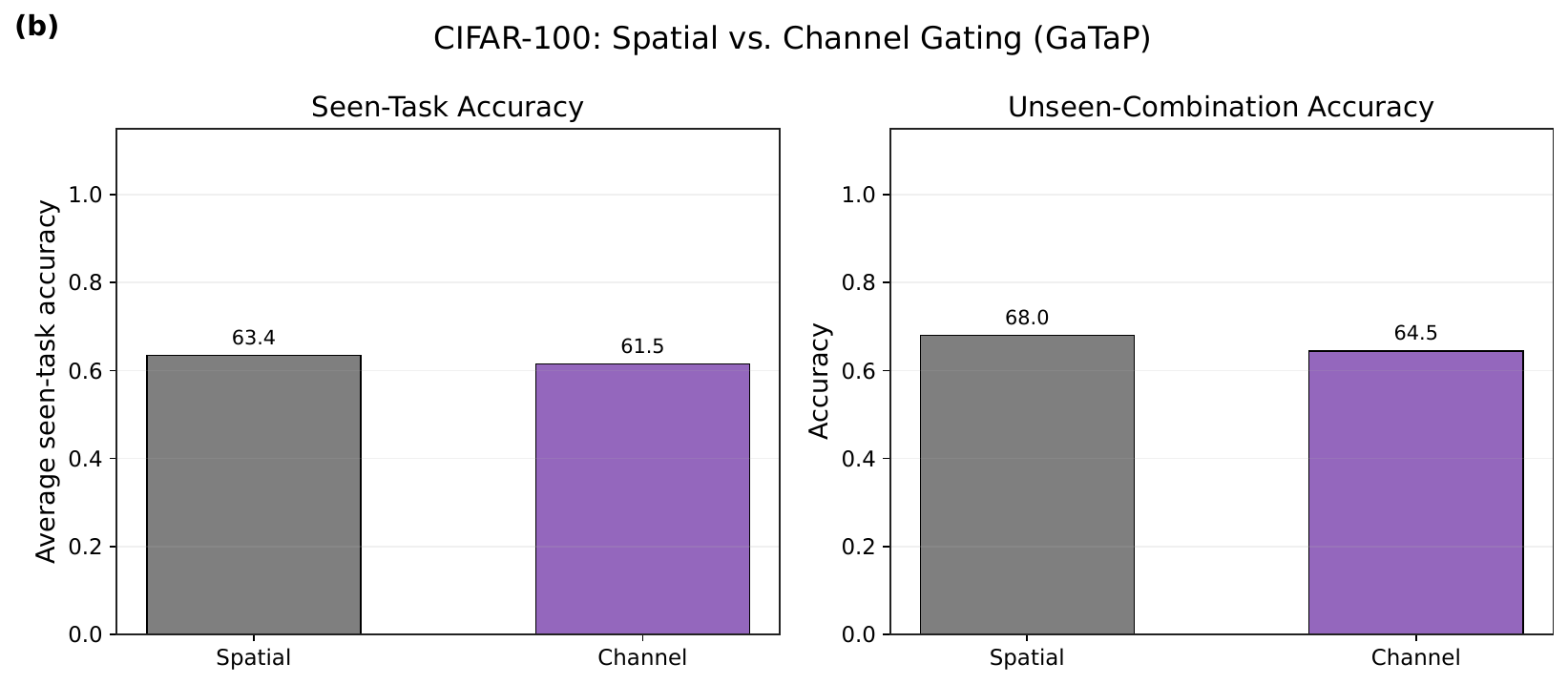}
    \caption{A comparison of seen- (left) and unseen-combination (right) task accuracy when using GaTaP in conjunction with spatial vs. channel gating for convolutional neural network models on \textbf{a)} CIFAR-10 and \textbf{b)} CIFAR-100.}
    \label{fig:spatial_vs_channel_gating}
\end{figure}

CNNs are needed to evaluate GaTaP on realistic image tasks, where spatial structure and local feature extraction are essential. Extending the framework to convolutions tests whether the same gating and target-inference mechanisms remain effective beyond fully connected networks. Gating in a convolutional layer could be accomplished by assigning one gate per output channel, applying the same gate to \textit{all spatial locations}, or a separate gate could be assigned for each channel at each spatial position (see Supplemental Figure \ref{fig:spatial-gating} for a schematic). We begin with the ``spatial gating'' version, before introducing the ``channel gating'' version. We find empirically that spatial gating is more effective (Supplemental Figure \ref{fig:spatial_vs_channel_gating}), and so we use this gating mechanism for our main results.

\begin{figure*}[t]
    \centering
    \begin{subfigure}[t]{0.49\textwidth}
        \centering
        \includegraphics[width=\linewidth]{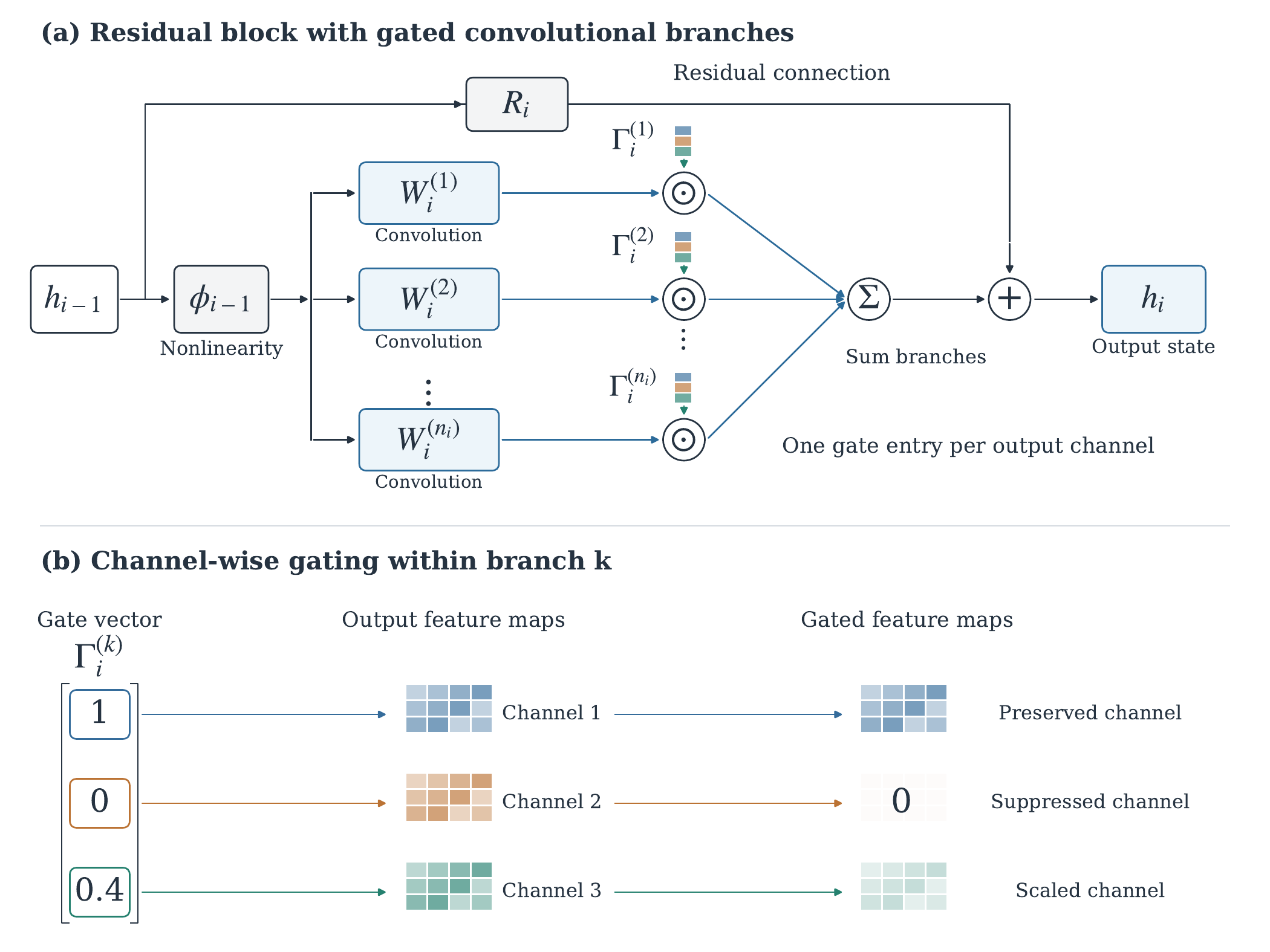}
        \caption{Channel-wise gating.}
        \label{fig:channel-gating}
    \end{subfigure}
    \hfill
    \begin{subfigure}[t]{0.49\textwidth}
        \centering
        \includegraphics[width=\linewidth]{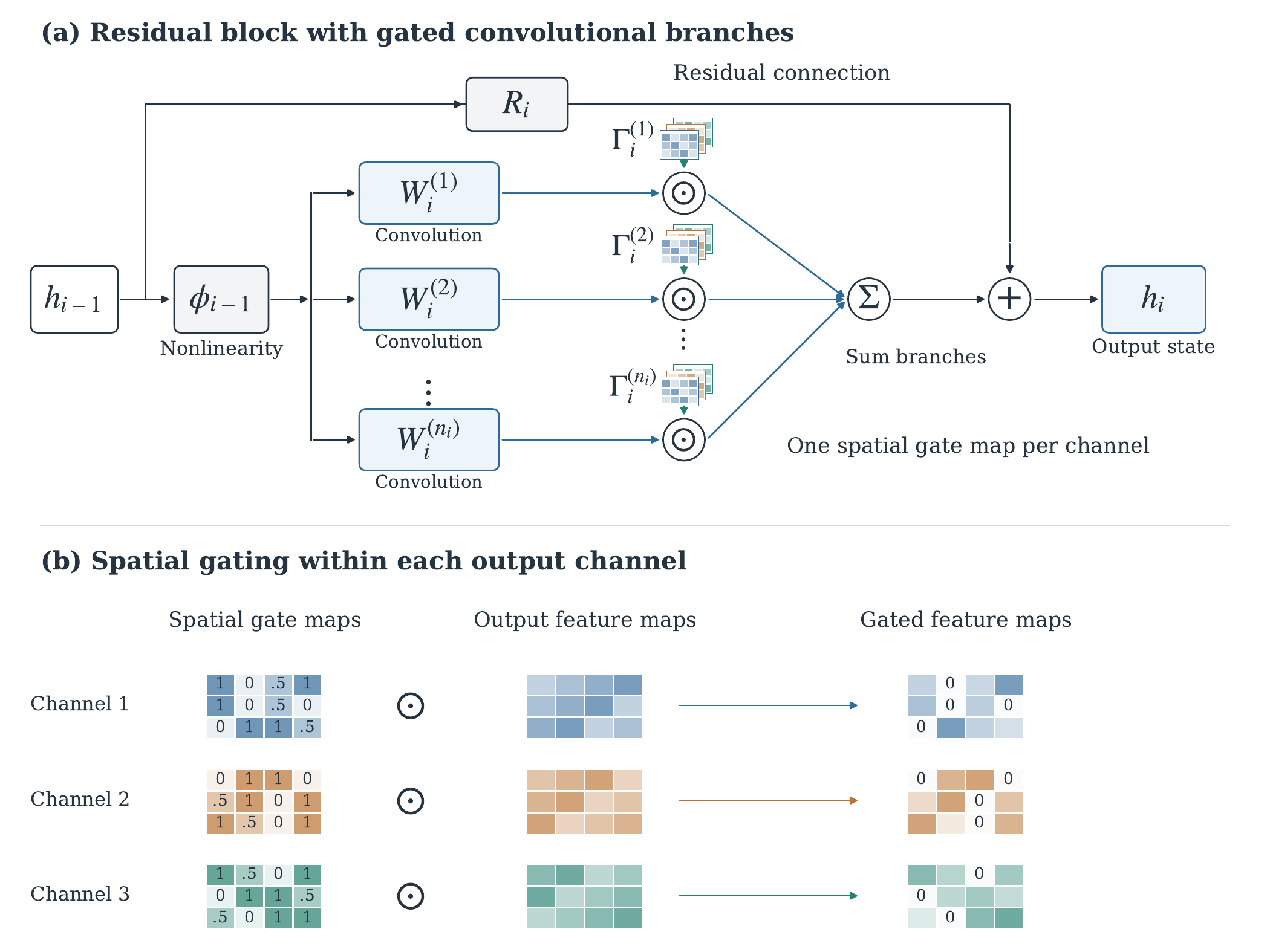}
        \caption{Spatial gating within each channel.}
        \label{fig:spatial-gating}
    \end{subfigure}

    \caption{
    \textbf{Two gating mechanisms for residual convolutional networks.}
    Both variants combine parallel convolutional branches with
    task-specific gates and add the residual contribution
    $R_i(h_{i-1})$ to the sum of the gated branch outputs.
    \textbf{Left:} Channel-wise gating assigns one scalar to each
    output channel, uniformly scaling its entire feature map.
    \textbf{Right:} Spatial gating assigns a separate gate value
    to each spatial location within each output channel, enabling
    selective modulation within a feature map.
    The lower panels illustrate how gate values preserve,
    suppress, or attenuate the corresponding activations.
    }
    \label{fig:cnn-gating-mechanisms}
\end{figure*}

\subsubsection{Case of multi subnetworks: spatial gating}
\label{CloseformsolutionCNNs_Multiplesubnetworks_perlayer_spatial_gating}

For a convolutional layer with multiple subnetworks, we would have kernels
\begin{align}
    W_{i}^{(k)} \in \mathbb{R}^{C_{i}\times I_{i} \times k_{h_i} \times k_{w_i}},
\end{align}
where $C_{i}$ is the feature output number of channel, $I_{i}$ is the input number of channel, $k_{h_i}, k_{w_i}$ are the height and width of the kernels.

We propose the following model for the convolutional layer: 
\begin{align}
    h_{i} = R_i \left( h_{i-1} \right) + \sum_{k=1}^{n_i} \Gamma_i^{(k)} \odot \left( W_i^{(k)} * \phi_{i-1}(h_{i-1}) \right) +\epsilon_i, \,\,\,\text{ for } \,i = 2, \ldots, L,
\end{align}
where
\begin{itemize}
    \item $n_i:$ number of kernels or shared knowledge between tasks,
    \item $\phi_{i-1}:$ is the nonlinearity function,
    \item $\Gamma_{i}^{(k)} \in \mathbb{R}^{C_{i}}$ plays the role of gating mechanism across the rows of $W_{i}^{(k)} \in \mathbb{R}^{C_{i}\times I_{i} \times k_{h_i} \times k_{w_i}}$. 
\end{itemize}

Let $a_{i-1,p} = \phi_{i-1}(h_{i-1, p})$ for an example $p$ in the support set $p=1,\ldots, P$. For subnet $k$, let $Z_{i,p}^{(k)} = W_{i}^{(k)} \ast a_{i-1, p}$. Assume a convolution layer has $C_{i}$ output channels, then $Z_{i,p}^{(k)} \in \mathbb{R}^{C_{i}\times H_{i} \times W_{i}}$. Let $u=(u_1,u_2)$ be the spatial position in the output feature map. So $Z_{i,p,c,u}^{(k)}$ is such that $p$ is the example index, $c$ is the output channel, $u$ is the spatial position and $k$ is the subnetwork index. $\Gamma_{i}^{(k)} \in \mathbb{R}^{C_{i} \times H_{i} \times W_{i}}$ so that $\left[ \Gamma_{i}^{(k)} Z_{i,p}^{(k)}\right]_{c, u} = \gamma_{i,c, u}^{(k)} Z_{i,p,c,u}^{(k)}$ and
\begin{align}
    h_{i,p,c,u} = \sum_{k=1}^{n_i}\gamma_{i,c,u}^{(k)} Z_{i,p,c,u}^{(k)}.
\end{align}
Let assume we have $h_{i,p,c,u}^{*}$ which is the target hidden state. We define the local loss as:
\begin{align}
    \mathcal{L}_{i} \left( \Gamma_{i}^{(k)} \right) = \dfrac{1}{2 P \lvert \Omega_{i}\rvert} \sum_{p=1}^{P} \sum_{c=1}^{C_{i}} \sum_{u\in \Omega_{i}} \left( h_{i,p,c,u}^{*} - R_i \left( h_{i-1} \right) - \sum_{k=1}^{n_i}\gamma_{i,c,u}^{(k)} Z_{i,p,c,u}^{(k)} - \epsilon_i \right)^{2} + \alpha_{i} \sum_{k=1}^{n_{i}} \sum_{c=1}^{C_{i}} \sum_{u\in \Omega_{i}} \left\lvert \gamma_{i,c,u}^{(k)}\right\rvert.
\end{align}
Let
\begin{align}
    R_{i,p,c,u}^{(-k)} = h_{i,p,c,u}^{*} - R_i \left( h_{i-1} \right) - \sum_{l\neq k}\gamma_{i,c,u}^{(l)} Z_{i,p,c,u}^{(l)} - \epsilon_i.
\end{align}
For $\gamma_{i,c,u}^{(k)}$, we have its corresponding loss given by:
\begin{align}
    \mathcal{L}_{i,k,c,u} \left( \gamma_{i,c}^{(k)} \right) = \dfrac{1}{2 P \lvert \Omega_{i}\rvert} \sum_{p=1}^{P}  \left( R_{i,p,c,u}^{(-k)} - \gamma_{i,c,u}^{(k)} Z_{i,p,c,u}^{(k)} \right)^{2} + \alpha_{i} \left\lvert \gamma_{i,c,u}^{(k)}\right\rvert.
\end{align}
With the following definitions:
\begin{align}
    b_{i,k,c,u} = \dfrac{1}{P \lvert \Omega_{i}\rvert} \sum_{p=1}^{P} Z_{i,p,c,u}^{(k)}R_{i,p,c,u}^{(-k)} \,\,\text{ and }\,\, a_{i,k,c,u} = \dfrac{1}{P \lvert \Omega_{i}\rvert} \sum_{p=1}^{P}  \left( Z_{i,p,c,u}^{(k)} \right)^{2},
\end{align}
the closed-form solution for $\gamma_{i,c,u}^{(k)}$ (directly following from \Eqref{eq:closed_form_gamma}) is given by:
\begin{align}
    \left( \gamma_{i,c,u}^{(k)}\right)^{*} = \text{sign}\left( b_{i,k,c,u} \right) \times \left( \dfrac{\left\lfloor \left\lvert b_{i,k,c,u}\right\rvert - \alpha_{i} \right\rfloor^{+}}{a_{i,k,c,u}}\right).
\end{align}

\subsubsection{Case of multi subnetworks: channel gating}
\label{CloseformsolutionCNNs_Multiplesubnetworks_perlayer_channel_gating}

Let $a_{i-1,p} = \phi_{i-1}(h_{i-1, p})$ for an example $p$ in the support set $p=1,\ldots, P$. For subnet $k$, let $Z_{i,p}^{(k)} = W_{i}^{(k)} \ast a_{i-1, p}$. Assume a convolution layer has $C_{i}$ output channels, then $Z_{i,p}^{(k)} \in \mathbb{R}^{C_{i}\times H_{i} \times W_{i}}$. So $Z_{i,p,c}^{(k)}$ is such that $p$ is the example index, $c$ is the output channel, and $k$ is the subnetwork index. $\Gamma_{i}^{(k)} \in \mathbb{R}^{C_{i}}$ so that $\left[ \Gamma_{i}^{(k)} Z_{i,p}^{(k)}\right]_{c} = \gamma_{i,c}^{(k)} Z_{i,p,c}^{(k)}$ and 
\begin{align}
    h_{i,p,c} = \sum_{k=1}^{n_i}\gamma_{i,c}^{(k)} Z_{i,p,c}^{(k)}.
\end{align}
Let assume we have $h_{i,p,c}^{*}$ which is the target hidden state. We define the local loss as:
\begin{align}
    \mathcal{L}_{i} \left( \Gamma_{i}^{(k)} \right) = \dfrac{1}{2 P \lvert \Omega_{i}\rvert} \sum_{p=1}^{P} \sum_{c=1}^{C_{i}} \left( h_{i,p,c}^{*} - R_i \left( h_{i-1} \right) - \sum_{k=1}^{n_i}\gamma_{i,c}^{(k)} Z_{i,p,c}^{(k)} - \epsilon_i \right)^{2} + \alpha_{i} \sum_{k=1}^{n_{i}} \sum_{c=1}^{C_{i}} \left\lvert \gamma_{i,c}^{(k)}\right\rvert.
\end{align}
Let
\begin{align}
    R_{i,p,c}^{(-k)} = h_{i,p,c}^{*} - R_i \left( h_{i-1} \right) - \sum_{l\neq k}\gamma_{i,c}^{(l)} Z_{i,p,c}^{(l)} - \epsilon_i.
\end{align}
For $\gamma_{i,c}^{(k)}$, we have its corresponding loss given by:
\begin{align}
    \mathcal{L}_{i,k,c} \left( \gamma_{i,c}^{(k)} \right) = \dfrac{1}{2 P \lvert \Omega_{i}\rvert} \sum_{p=1}^{P}  \left( R_{i,p,c}^{(-k)} - \gamma_{i,c}^{(k)} Z_{i,p,c}^{(k)} \right)^{2} + \alpha_{i} \left\lvert \gamma_{i,c}^{(k)}\right\rvert.
\end{align}
With the following definitions:
\begin{align}
    b_{i,k,c} = \dfrac{1}{P \lvert \Omega_{i}\rvert} \sum_{p=1}^{P} Z_{i,p,c}^{(k)}R_{i,p,c}^{(-k)} \,\,\text{ and }\,\, a_{i,k,c} = \dfrac{1}{P \lvert \Omega_{i}\rvert} \sum_{p=1}^{P}  \left( Z_{i,p,c}^{(k)} \right)^{2},
\end{align}
the closed-form solution for $\gamma_{i,c}^{(k)}$ (again following from \Eqref{eq:closed_form_gamma}) is given by
\begin{align}
    \left( \gamma_{i,c}^{(k)}\right)^{*} = \text{sign}\left( b_{i,k,c} \right) \times \left( \dfrac{\left\lfloor \left\lvert b_{i,k,c}\right\rvert - \alpha_{i} \right\rfloor^{+}}{a_{i,k,c}}\right).
\end{align}

\subsubsection{Target hidden states}
\label{targetupdateCNNS}
Let
\begin{align}
    h_{i} = F_{i} \left( h_{i-1}\right) = R_i \left( h_{i-1} \right) + \sum_{k=1}^{n_i} \Gamma_{i}^{(k)} \odot \left(  W_{i}^{(k)}   \ast \phi_{i}\left(h_{i-1}\right)\right) + \epsilon_i
\end{align}
The same target update rule, as in Eq. \ref{targetupdate}, is used in the convolutional case. Let $\Sigma_{i-1} = \sigma_{i-1}^{2} I$, $\Sigma_{i} = \sigma_{i}^{2} I$, and $e_{i} = h_{i}^{*} - h_{i}$. Therefore 
\begin{align}
    h_{i-1}^{*} &= h_{i-1} + \sigma_{i-1}^{2} J_{i}^{T} \left( \sigma_{i-1}^{2} J_{i}  J_{i}^{T} + \sigma_{i}^{2} I\right)^{-1} e_{i}\\
    &= h_{i-1} + \sigma_{i-1}^{2} J_{i}^{T} \left( \sigma_{i-1}^{2} J_{i}  J_{i}^{T} + \sigma_{i}^{2} I\right)^{-1} e_{i}
\end{align}
For large convolutional networks, the Kalman gain $K_i$ is computationally very expensive because of the large inverse calculation $\left( \sigma_{i-1}^{2} J_{i}  J_{i}^{T} + \sigma_{i}^{2} I\right)^{-1}$. We therefore use a diagonal approximation in our convolutional experiments that renders inversion trivial.

\paragraph{\textbf{Use of the diagonal of $J_i J_i^T$}:}
Here, for convolutional neural network models, we simply take a diagonal approximation of the $J_i J_i^T$ matrix in calculating the Kalman gain as follows:
\begin{align}
    h_{i-1}^{*} &= h_{i-1} + \sigma_{i-1}^{2} J_{i}^{T} \left( \sigma_{i-1}^{2} J_{i}  J_{i}^{T} + \sigma_{i}^{2} I\right)^{-1} e_{i}\\
    &\approx h_{i-1} + \sigma_{i-1}^{2} J_{i}^{T} \left( \sigma_{i-1}^{2} \text{diag} \left( J_{i}  J_{i}^{T} \right) + \sigma_{i}^{2} I\right)^{-1} e_{i}\\
    &= h_{i-1} + \sigma_{i-1}^{2} J_{i}^{T} \left[ \text{diag} \left( \sigma_{i-1}^{2}  J_{i}  J_{i}^{T}  + \sigma_{i}^{2} I\right)\right]^{-1} e_{i}.
\end{align}
Since inverting a diagonal matrix is equivalent to pointwise inversion of each diagonal matrix element, this makes computing the inverse much simpler. In practice, we find that it does not significantly impact training performance.

\begin{algorithm}[h]
\caption{Gated Target Propagation}
\label{alg:gated_targetprop}
\begin{algorithmic}[1]

\STATE \textbf{Training phase}
\STATE Initialize all shared subnetworks $\{W_i^{(k)}\}$

\FOR{$j = 1,\ldots,m$}

    \STATE Receive task $\mathcal{T}_j$
    \STATE Split task data into support set $\mathcal{S}_j$
    and query set $\mathcal{Q}_j$

    \FOR{$e = 1,\ldots,E$}

        \STATE \textbf{Inner loop: infer task-specific gates}

        \STATE Forward $\mathcal{S}_j$ through the network
        \STATE Compute the output target $h_L^\star$ followed by $\Gamma_{j,L}^\star$

        \FOR{$i = L-1,\ldots,1$}
            \STATE Compute target state $h_i^\star$
            using target propagation
            \STATE Infer gamma parameters $\Gamma_{j,i}^\star$
            from $\mathcal{S}_j$
            using the closed-form sparse update
        \ENDFOR
        
        \STATE \textbf{Outer loop: update shared weights}
        \STATE Keep $\Gamma_j^\star$ fixed
        \STATE Forward $\mathcal{Q}_j$ through the gated network
        \STATE Compute query target states
        $\{h_i^\star\}_{i=1}^{L}$

        \STATE Compute the local target-propagation loss $\mathcal{L}_{i,\mathrm{query}}$
        \STATE Update only the shared weights

        \STATE
        $\displaystyle
        W_i^{(k)}
        \leftarrow
        W_i^{(k)}
        -
        \eta
        \nabla_{W_i^{(k)}}
        \mathcal{L}_{i,\mathrm{query}}
        $

    \ENDFOR

\ENDFOR

\STATE \textbf{Testing phase}
\STATE Freeze all shared subnetworks $\{W_i^{(k)}\}$

\FOR{each evaluation task $\mathcal{T}$}

    \STATE Split task data into support set $\mathcal{S}$
    and query set $\mathcal{Q}$

    \STATE Infer gamma parameters
    $\Gamma_{\mathcal{T}}^\star$
    from $\mathcal{S}$ only

    \STATE Evaluate loss and accuracy on $\mathcal{Q}$

\ENDFOR

\end{algorithmic}
\end{algorithm}

\begin{figure}[p]
    \centering
    \includegraphics[width=\linewidth,height=0.3\textheight,keepaspectratio]{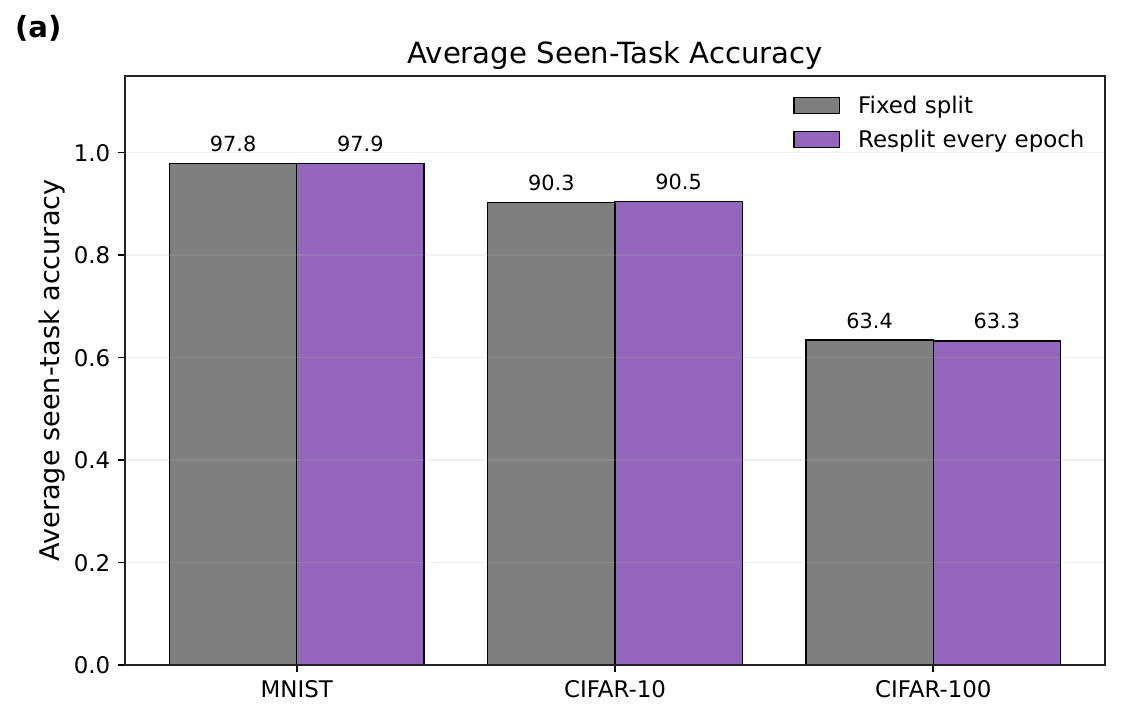}\\[4pt]
    \includegraphics[width=\linewidth,height=0.3\textheight,keepaspectratio]{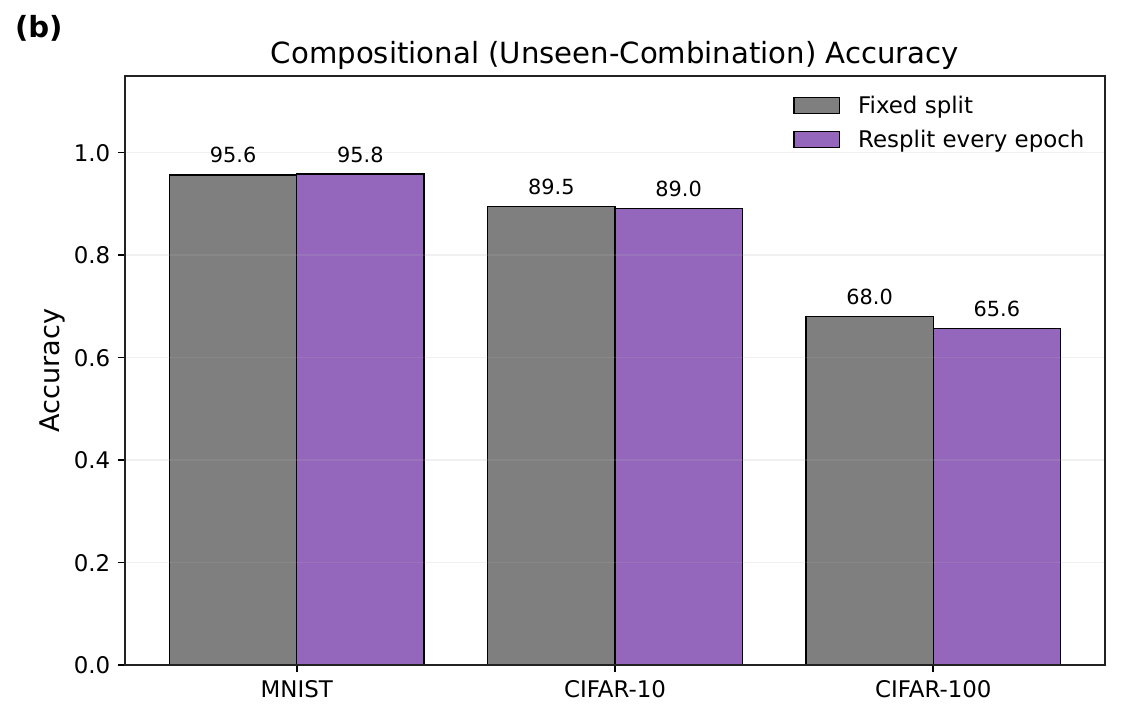}\\[4pt]
    \includegraphics[width=\linewidth,height=0.3\textheight,keepaspectratio]{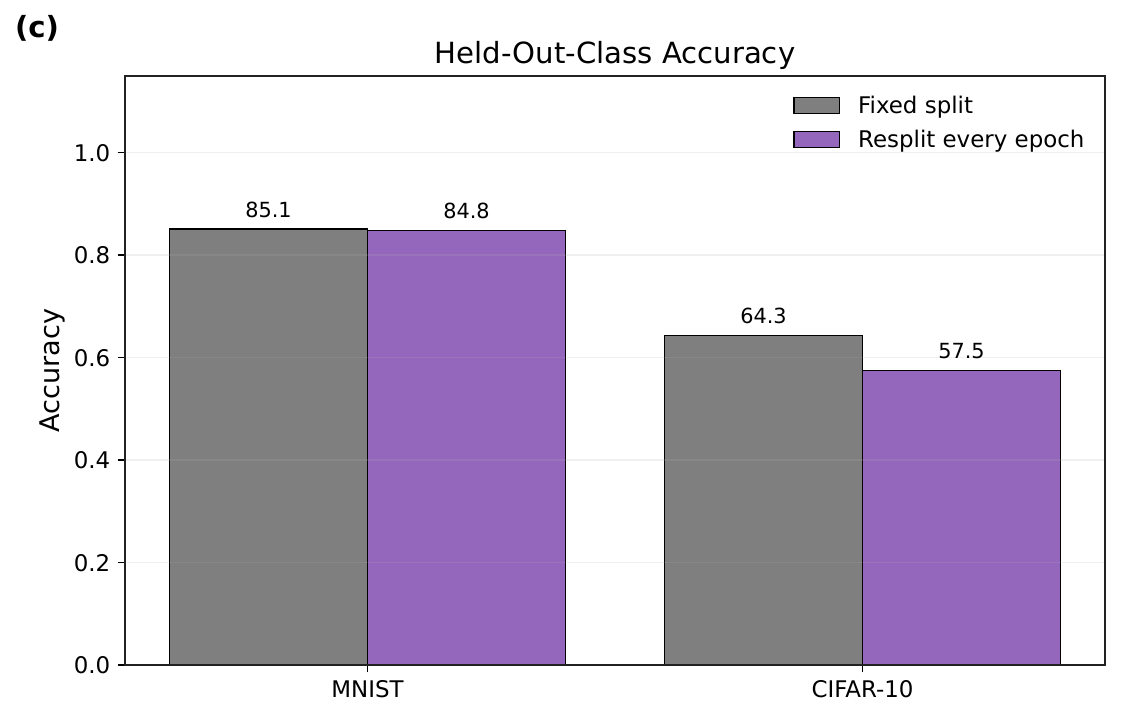}
    \caption{Resplitting the support and query sets every epoch, as compared to a fixed split used across all epochs, as evaluated by \textbf{a)} average seen-task accuracy, \textbf{b)} compositional (unseen-combination) accuracy, and \textbf{c)} held-out-class accuracy.}
    \label{fig:resplit_vs_fixed_split}
\end{figure}

\begin{figure}[p]
    \centering
    \includegraphics[width=\linewidth]{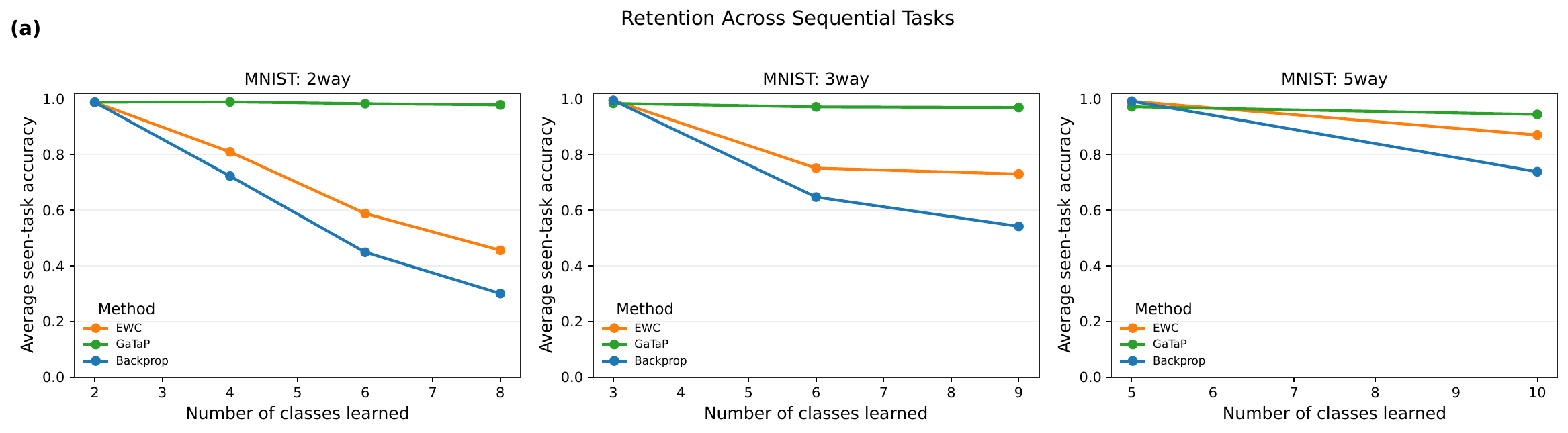}\\[4pt]
    \includegraphics[width=\linewidth]{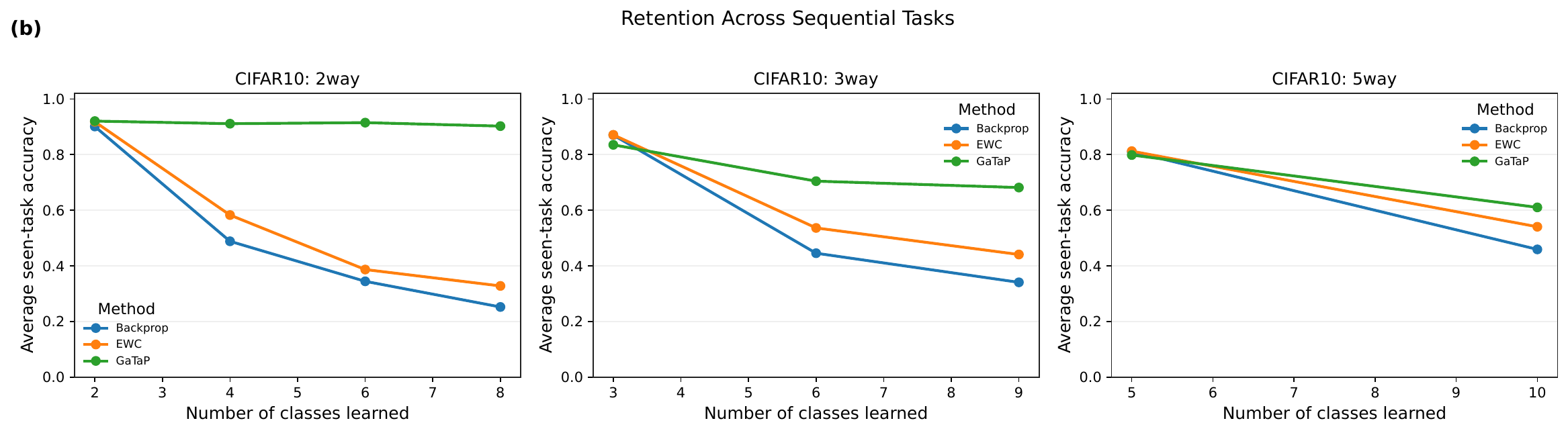}\\[4pt]
    \includegraphics[width=\linewidth]{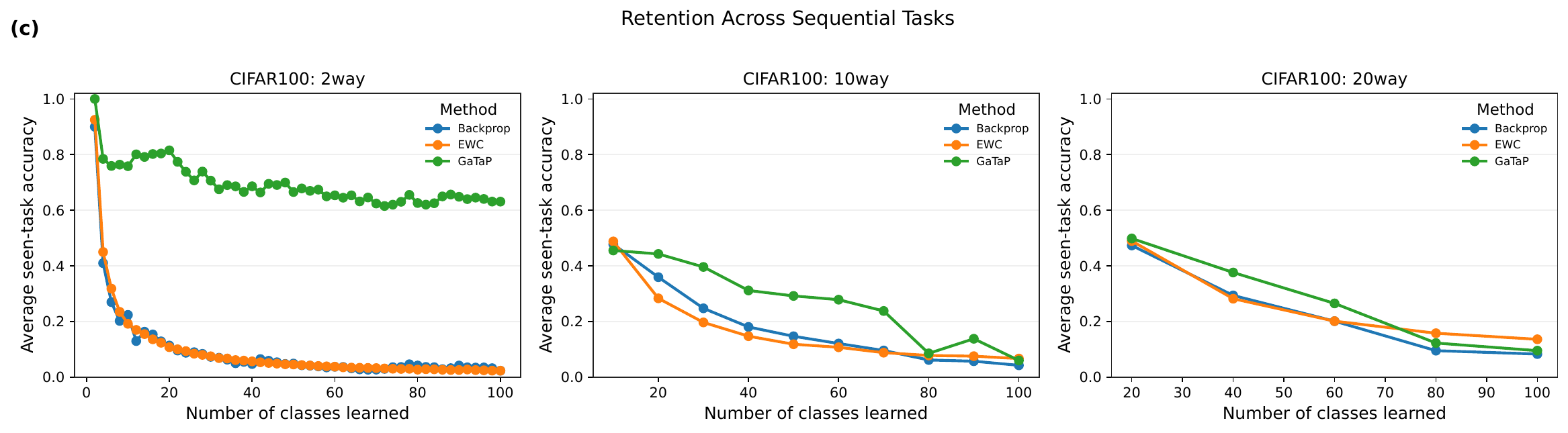}
    \caption{Retention across sequential tasks with varying numbers of classes per-task. Networks are trained with GaTaP (green), EWC (orange), and standard backpropagation (blue). \textbf{a)} Models trained on MNIST, for two-way, three-way and five-way continual learning classification tasks. \textbf{b)} Models trained on CIFAR-10, for two-way, three-way, and five-way continual learning classification tasks. \textbf{c)} Models trained on CIFAR-100, for two-way, ten-way, and twenty-way continual learning classification tasks.}
    \label{fig:retention_across_tasks}
\end{figure}

\newpage

\subsection{Hyperparameters}
\label{Hyperparameters}

\begin{table}[h]
\centering
\caption{Shared training hyperparameters for MNIST (2-way, 3-way, and 5-way protocols). Values are identical across all three protocols and all three methods (GaTaP, EWC, Backprop) except where noted.}
\label{tab:mnist_hyperparams_shared}
\begin{tabular}{ll}
\toprule
Hyperparameter & Value \\
\midrule
Architecture & Feedforward MLP \\
Hidden layer sizes & 64, 128 \\
Epochs per task & 10 \\
Learning rate ($lr_w$) & $1 \times 10^{-3}$ \\
Optimizer & Adam \\
Random seed & 42 \\
\bottomrule
\end{tabular}
\end{table}

\begin{table}[h]
\centering
\caption{Method-specific hyperparameters for MNIST. GaTaP columns apply identically across the 2-way, 3-way, and 5-way protocols.}
\label{tab:mnist_hyperparams_method}
\begin{tabular}{ll}
\toprule
\multicolumn{2}{l}{\textbf{GaTaP}} \\
\midrule
Number of subnetworks & 1 \\
Sparsity coefficient ($\alpha$) & $10^{-5}$ \\
Gamma inference steps (train / test) & 10 / 10 \\
Noise scale by layer ($\sigma$) & 0.1, 0.1, 0.1 \\
Support examples per class (train / test) & 32 / 64 \\
Query batch size (train / test) & 128 / 256 \\
\midrule
\multicolumn{2}{l}{\textbf{EWC}} \\
\midrule
EWC regularization strength ($\lambda$) & 20{,}000 \\
Fisher information sample size & 1{,}000 \\
Fisher information batch size & 32 \\
\bottomrule
\end{tabular}
\end{table}

\begin{table}[h]
\centering
\caption{Shared training hyperparameters for CIFAR-10 (2-way, 3-way, and 5-way protocols). Values are identical across all three protocols and all three methods (GaTaP, EWC, Backprop) except where noted.}
\label{tab:cifar10_hyperparams_shared}
\begin{tabular}{ll}
\toprule
Hyperparameter & Value \\
\midrule
Architecture & Simple ResNet (CNN) \\
Channel widths & 16, 32, 64, 128, 256, 512 \\
Kernel size & 5 \\
Epochs per task & 10 \\
Learning rate ($lr_w$) & $1 \times 10^{-3}$ \\
Optimizer & Adam \\
Random seed & 42 \\
\bottomrule
\end{tabular}
\end{table}

\begin{table}[h]
\centering
\caption{Method-specific hyperparameters for CIFAR-10. GaTaP columns apply identically across the 2-way, 3-way, and 5-way protocols.}
\label{tab:cifar10_hyperparams_method}
\begin{tabular}{ll}
\toprule
\multicolumn{2}{l}{\textbf{GaTaP}} \\
\midrule
Number of subnetworks & 1 \\
Gamma mode & Spatial \\
Sparsity coefficient ($\alpha$) & $10^{-5}$ \\
Gamma scale & 1.0 \\
Gamma inference steps (train / test) & 10 / 10 \\
Noise scale by layer ($\sigma$) & 0.1 (all 7 layers) \\
Support examples per class (train / test) & 20 / 64 \\
Query batch size (train / test) & 128 / 256 \\
\midrule
\multicolumn{2}{l}{\textbf{EWC}} \\
\midrule
EWC regularization strength ($\lambda$) & 5{,}000 \\
Fisher information sample size & 1{,}000 \\
Fisher information batch size & 32 \\
\bottomrule
\end{tabular}
\end{table}

\begin{table}[h]
\centering
\caption{Shared training hyperparameters for CIFAR-100 (2-way, 10-way, and 20-way protocols). Values are identical across all three protocols and all three methods (GaTaP, EWC, Backprop) except where noted.}
\label{tab:cifar100_hyperparams_shared}
\begin{tabular}{ll}
\toprule
Hyperparameter & Value \\
\midrule
Architecture & Simple ResNet (CNN) \\
Channel widths & 16, 32, 64, 128, 256, 512 \\
Kernel size & 5 \\
Epochs per task & 20 \\
Learning rate ($lr_w$) & $3 \times 10^{-4}$ \\
Optimizer & Adam \\
Random seed & 42 \\
\bottomrule
\end{tabular}
\end{table}

\begin{table}[h]
\centering
\caption{Method-specific hyperparameters for CIFAR-100. GaTaP columns apply identically across the 2-way, 10-way, and 20-way protocols.}
\label{tab:cifar100_hyperparams_method}
\begin{tabular}{ll}
\toprule
\multicolumn{2}{l}{\textbf{GaTaP}} \\
\midrule
Number of subnetworks & 2 \\
Gamma mode & Spatial \\
Sparsity coefficient ($\alpha$) & $10^{-7}$ \\
Gamma scale & 1.0 \\
Gamma inference steps (train / test) & 10 / 10 \\
Noise scale by layer ($\sigma$) & 0.1 (all 7 layers) \\
Support examples per class (train / test) & 20 / 64 \\
Query batch size (train / test) & 128 / 256 \\
\midrule
\multicolumn{2}{l}{\textbf{EWC}} \\
\midrule
EWC regularization strength ($\lambda$) & 100{,}000 \\
Fisher information sample size & 1{,}000 \\
Fisher information batch size & 32 \\
\bottomrule
\end{tabular}
\end{table}

\end{document}

%% file: math_commands.tex
\usepackage{amsmath,amsfonts,bm}

\def\eqref#1{equation~\ref{#1}}
\def\Eqref#1{Equation~\ref{#1}}
\def\1{\bm{1}}

\DeclareMathAlphabet{\mathsfit}{\encodingdefault}{\sfdefault}{m}{sl}
\SetMathAlphabet{\mathsfit}{bold}{\encodingdefault}{\sfdefault}{bx}{n}